\documentclass{article}

\usepackage[main, final]{neurips_2026}

\usepackage[utf8]{inputenc}
\usepackage[T1]{fontenc}
\usepackage{booktabs}
\usepackage{amsfonts}
\usepackage{amsmath}
\usepackage{amssymb}
\usepackage{nicefrac}
\usepackage{microtype}
\usepackage[table]{xcolor}
\usepackage{graphicx}
\usepackage{makecell}
\usepackage{multirow}
\usepackage{enumitem}
\usepackage{algorithm}
\usepackage{algpseudocode}
\usepackage{placeins}
\usepackage{float}
\usepackage{tikz}
\usetikzlibrary{arrows.meta,positioning,fit}
\usepackage{url}
\usepackage{wrapfig}
\usepackage{fvextra}
\usepackage{fancyvrb}

\usepackage[pagebackref=true,breaklinks=true,colorlinks=true,bookmarks=false]{hyperref}

\definecolor{deepred}{HTML}{940000}

\hypersetup{linkcolor=deepred}
\hypersetup{citecolor=[rgb]{0.4,0.15,0.95}}

\usepackage{tocloft}
\usepackage[toc,page,header]{appendix}
\usepackage{minitoc}

\renewcommand \thepart{}
\renewcommand \partname{}

\usepackage[T1]{fontenc}    % use 8-bit T1 fonts
\usepackage{caption}
\newcommand{\DPOone}{\textsc{DPO}\ensuremath{_{\text{1-step}}}}
\newcommand{\GRPOone}{\textsc{GRPO}\ensuremath{_{\text{1-step}}}}
\newcommand{\DPOonemath}{\mathrm{DPO}_{\text{1-step}}}

\newcommand{\DRaFT}{DRaFT}
\newcommand{\DrPO}{DrPO}

\newcommand{\PSO}{\textsc{PSO}}
\newcommand{\VGGFlow}{\ensuremath{\mathrm{VGG\mbox{-}Flow}_{\text{1-step}}}}

\title{Drifting Preference Optimization for One-Step Generative Models}

\author{%
Zhou Jiang$^{1}$ \quad Yandong Wen$^{2}$ \quad Zhen Liu$^{1}$\thanks{Corresponding Author.} \\
$^{1}$The Chinese University of Hong Kong, Shenzhen \quad $^{2}$Westlake University 
}

\begin{document}

\doparttoc % Tell to minitoc to generate a toc for the parts
\faketableofcontents

\maketitle

\begin{abstract}
% One-step text-to-image generators are attractive for deployment because they produce an image with a single forward pass, but their preference finetuning remains difficult: standard alignment methods often require policy likelihoods, denoising trajectories, differentiable reward gradients, or test-time optimization. We propose Drifting Preference Optimization (\DrPO{}), an online preference-finetuning method for deterministic one-step generators. \DrPO{} uses the target reward only to rank on-policy samples, constructs positive and negative supports for a non-parametric dipole reward model, and turns the resulting local preference field into a detached feature-space regression target. A reference drift estimated from the frozen base generator stabilizes the update, while inference remains a single generator call. We evaluate \DrPO{} on SD-Turbo and SDXL-Turbo with multiple target rewards and benchmarks, including HPSv3 and GenEval. Across these settings, \DrPO{} improves preference alignment over one-step preference baselines, avoids backpropagation through large or non-differentiable rewards, and achieves a $3.51\times$ per-update speedup over reward-gradient finetuning with HPSv3 under the matched effective batch setting. Preliminary offline experiments further suggest that sample-based gradient synthesis can be extended beyond online reward ranking.
One-step text-to-image generators are attractive for deployment because they generate an image with a single forward pass, but preference finetuning them remains difficult: standard alignment methods often rely on policy likelihoods, denoising trajectories, differentiable reward gradients, or test-time optimization. We propose Drifting Preference Optimization (\DrPO{}), an online preference-finetuning method for deterministic one-step generators. For each prompt, \DrPO{} samples candidates from the current generator, ranks them with a target reward, and uses high- and low-scoring samples to synthesize a feature-space update direction. The update is a non-parametric population-rank preference field plus a reference drift estimated from the frozen base generator, and is optimized through a detached feature-space regression target. The target reward is used only for ranking, so \DrPO{} can train with large, black-box, or non-differentiable rewards while inference remains a single generator call. We evaluate \DrPO{} on SD-Turbo and SDXL-Turbo with multiple target rewards and benchmarks, including HPSv3 and GenEval. \DrPO{} improves alignment over reward-gradient-free one-step preference baselines and reduces HPSv3 training computation by $3.51\times$ under the matched effective-batch setting by removing reward-model backpropagation. Initial offline experiments suggest that sample-based gradient synthesis can also be used beyond online reward ranking. The project is available at \href{https://ugvly.github.io/DrPO/}{https://ugvly.github.io/DrPO/}.
\end{abstract}

\section{Introduction}
\label{sec:intro}

% One-step text-to-image generators have become an increasingly practical class of visual generative models, driven by progress in one-step generative training ~\citep{deng2026drifting,geng2025mean,sauer2023stylegan} and diffusion distillation ~\citep{song2023consistency,luo2023lcm,sauer2024add,yin2024one}. Compared with multi-step diffusion samplers, these models generate an image with a single forward pass and are therefore attractive for low-latency generation. Beyond fast sampling, however, deployment also requires these generators to produce images that better match human judgments and task-specific criteria. Following the success of reinforcement learning from human feedback (RLHF) in language and diffusion models, it is natural to post-train one-step generators with reward models so that their outputs are better aligned with such preferences.
Recent text-to-image systems increasingly target low-latency generation through one-step generative training~\citep{deng2026drifting,geng2025mean,sauer2023stylegan} and diffusion distillation~\citep{song2023consistency,luo2023lcm,sauer2024add,yin2024one}. Compared with multi-step diffusion samplers, these models generate an image with a single forward pass, making them useful for interactive and resource-constrained deployment. Deployment, however, also requires generators to produce images that better match human preferences and task-specific criteria. Following the success of reinforcement learning from human feedback (RLHF) in language and diffusion models, it is natural to post-train one-step generators with reward models so that their outputs better align with such preferences.

% However, reward-based post-training is nontrivial for one-step generators. The model maps noise to an image through a single deterministic transition, whereas traditional RL methods rely on stochastic actions and policy probabilities to estimate how rewards should update the policy. Directly backpropagating the reward through the generated image avoids this difficulty and can be efficient for lightweight differentiable rewards. Yet this update follows the local gradient of the reward model, rather than a sample-level preference update under reference control, and can therefore overfit reward idiosyncrasies or move the generator away from the reliable region of the base model. It also requires differentiating through the reward model, which can be costly for large VLM-based rewards and is inapplicable to non-differentiable evaluators. This motivates recovering a reward-driven, reference-aware update direction from samples and reward evaluations alone.
Reward-based post-training is nontrivial for one-step generators. Standard RLHF and preference-optimization methods typically rely on stochastic policy probabilities, density ratios, or denoising trajectories to construct policy updates. Directly backpropagating a differentiable reward through the generated image avoids these requirements and can be effective for lightweight rewards, but it follows the local gradient of the reward model and requires differentiating through that reward. This becomes costly for large VLM-based rewards and impossible for black-box or discrete evaluators. We therefore seek a reward-driven, reference-aware update direction that can be recovered from samples and reward evaluations alone.

% We take inspiration from Drifting models~\citep{deng2026drifting}, which learn one-step generators by constructing a feature-space drifting field that moves generated samples toward target supports and away from negative supports. In preference finetuning, reward-ranked samples naturally define such supports: high-scoring candidates provide attractive anchors and low-scoring candidates provide repulsive anchors. A frozen reference generator supplies a second drift field that counteracts uncontrolled distribution shift. This idea leads to Drifting Preference Optimization, dubbed \DrPO{}, an efficient finetuning framework for one-step generators. Since \DrPO{} replaces target-reward gradients with sample-based preference drifts, it supports expensive, black-box, or discrete rewards while preserving one-step inference. We empirically evaluate \DrPO{} across multiple reward models and one-step generators built on SD1.5 and SDXL. On HPSv3 and GenEval, \DrPO{} improves preference alignment without differentiating through the target reward. We further conduct ablations on candidate count, feature geometry, kernel choice, velocity scale, and reference regularization, and provide preliminary evidence that the same sample-based gradient synthesis can be extended to offline preference finetuning.
We take inspiration from Drifting models~\citep{deng2026drifting}, which train one-step generators by constructing feature-space drift fields from finite batches of samples. In preference finetuning, reward-ranked candidates naturally define such a field: high-scoring samples provide attractive fields, while low-scoring samples provide repulsive fields. \DrPO{} turns these fields into a non-parametric population-rank preference field and independently estimates a reference drift from samples of the frozen base generator. The combined field gives a detached feature-space target for updating the current generator. Since the target reward is used only to rank generated candidates, \DrPO{} supports expensive, black-box, or discrete rewards while preserving one-step inference. We evaluate \DrPO{} across multiple reward models and one-step generators built on SD1.5 and SDXL. On HPSv3 and GenEval, \DrPO{} improves preference alignment without differentiating through the target reward. We also report ablations on candidate count, feature choice, kernel choice, velocity scale, and reference regularization, along with initial results on offline preference finetuning.

% \begin{figure*}[t]
%   \centering
%   \begin{minipage}[t]{0.575\textwidth}
%     \centering
%     {\small\bfseries Online \DrPO{} for One-step generation}\par\vspace{0.25ex}
%     \includegraphics[width=\linewidth]{figures/assets/alg1_v2_compressed.pdf}
%   \end{minipage}\hspace{0.8em}
%   \begin{tikzpicture}[baseline={(separator.north)}]
%     \coordinate (separator) at (0,0);
%     \draw[densely dashed, line width=0.6pt, gray] (0,0) -- (0,-2.12in);
%   \end{tikzpicture}\hspace{0.8em}
%   \begin{minipage}[t]{0.365\textwidth}
%     \centering
%     {\small\bfseries Training trajectory}\par\vspace{0.25ex}
%     \includegraphics[width=\linewidth]{figures/outputs/intro_trajectory.pdf}
%   \end{minipage}
%   \caption{
%       \textbf{Overview of online \DrPO{} for one-step generation.} Left: the current generator samples candidate images for each prompt, and the target reward ranks them into preferred and rejected pairs. A frozen feature extractor maps these images into latent feature space, where preference pairs define the dipole drift and samples from the frozen base generator provide the reference correction. The combined drift gives a detached target for training the generator. Right: qualitative samples from fixed test prompts along the online finetuning trajectory.
%   }
%   \label{fig:intro-teaser}
% \end{figure*}

\begin{figure}[t]
  \centering
  \includegraphics[width=\linewidth]{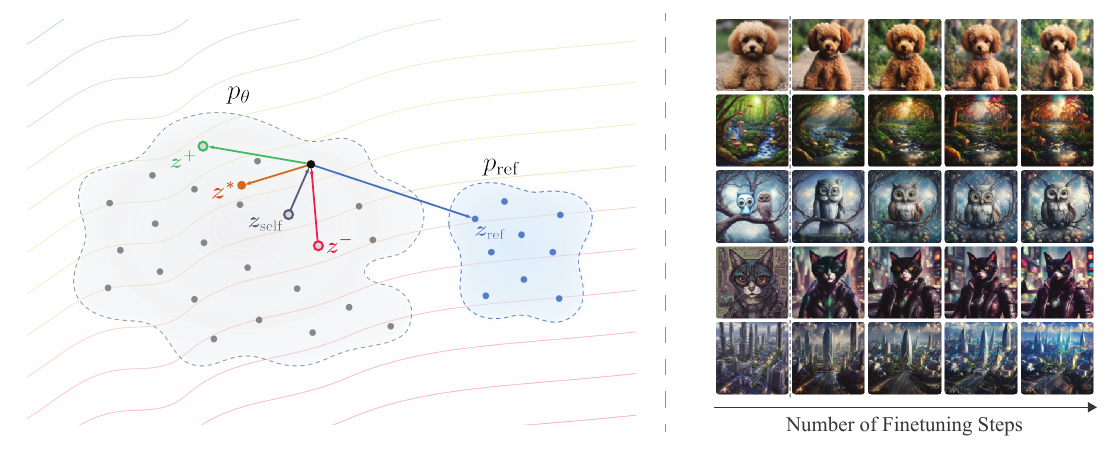}
  \caption{
      % \textbf{Overview of \DrPO{}.} Left: 
      % the construction of the drift field used to finetune the network. For an on-policy sample (black dot), $N$ random pairs are collected, in each of which the one with high reward labeled as positive (red dot) and the other labeled as negative (red dot). Together with the self-repulsion and attraction forces from $p_\theta$ and $p_\text{ref}$, the final regression target $z^*$ (brown dot) is synthesized. The contours represent the magnitude of reward values.
      % Right: qualitative samples from fixed test prompts along the online finetuning trajectory.
      \textbf{Overview of \DrPO{}.} Left: construction of the drift field used to finetune the network. For an on-policy sample (black dot), $N$ random pairs are collected. In each pair, the sample with the higher reward is labeled as positive (red dot), while the other is labeled as negative (blue dot). Together with the self-repulsion and attraction forces induced by $p_\theta$ and $p_\text{ref}$, respectively, these components synthesize the final regression target $z^*$ (brown dot). The contours indicate reward magnitudes. Right: qualitative samples from fixed test prompts along the online finetuning trajectory.
  }
  \label{fig:intro-teaser}
\end{figure}

Our core contributions are:
\begin{itemize}[leftmargin=*,nosep]
\setlength\itemsep{0.39em}
% \item We propose \DrPO{}, an online preference-finetuning method that turns reward-ranked on-policy samples into a detached drifting update for one-step generators while keeping inference to one forward pass.

% \item We introduce a dipole reward model that builds latent-space reward-gradient proxies from positive and negative supports, allowing expensive target-reward gradients to be replaced by sample-based drifting updates.

% \item We validate \DrPO{} on SD-Turbo and SDXL-Turbo across multiple rewards, show consistent gains over one-step preference baselines, and demonstrate a $3.51\times$ matched-batch update speedup over reward-gradient finetuning with HPSv3.

% \item We show that \DrPO{} works with non-differentiable GenEval-style rewards and provide preliminary results on offline preference finetuning.
\item We propose \DrPO{}, an online preference-finetuning method that converts reward-ranked on-policy samples into a detached drifting update for one-step generators while keeping inference to one forward pass.

\item We introduce a population-rank preference field that builds latent-space update directions from positive and negative samples, replacing target-reward gradients with sample-based drifting updates.

\item We validate \DrPO{} on SD-Turbo and SDXL-Turbo across multiple rewards, where it gives the largest gains among reward-gradient-free one-step preference baselines.

\item We show that \DrPO{} works with large and non-differentiable rewards, including HPSv3 and GenEval-style rewards, and provide preliminary results on offline preference finetuning.
\end{itemize}

\section{Related Work}
\label{sec:related_work}

\paragraph{One-step generative models.}
% Fast image generators often arise by compressing iterative sampling into one or a few evaluations through diffusion distillation~\citep{yin2024one, yin2024improved}, consistency training~\citep{lu2024simplifying}, adversarial objectives, flow distillation~\citep{nguyen2024swiftbrush}, or related mechanisms. This deployment form is attractive because sampling is a single forward pass, but it also removes many objects that standard alignment methods rely on: denoising trajectories, timestep-wise scores, sampler likelihoods, or easy policy log probabilities. Existing fast-diffusion alignment work often keeps some of that structure available during training, for example through diffusion-specific score regularization, pairwise margins, or step-level win-lose construction ~\citep{luo2024diffinstructpp,luo2024diffinstructstar,miao2024pso,croitoru2025curriculumdpo, luo2025rewardinstruct,luo2025jdm}. 
Fast image generators often compress iterative sampling into one or a few network evaluations through diffusion distillation~\citep{yin2024one,yin2024improved}, consistency training~\citep{lu2024simplifying}, adversarial objectives, flow distillation~\citep{nguyen2024swiftbrush}, or related mechanisms. Sampling then requires only a single forward pass, but many structures used by standard alignment methods are no longer available, including denoising trajectories, timestep-wise scores, sampler likelihoods, and tractable policy log probabilities. Existing fast-diffusion alignment methods often retain some diffusion-specific structure during training, for example through score regularization, pairwise margins, or step-level win--lose construction~\citep{luo2024diffinstructpp,luo2024diffinstructstar,miao2024pso,croitoru2025curriculumdpo,luo2025rewardinstruct,luo2025jdm}.

% \paragraph{General alignment strategies.}
% RLHF aligns generative policies by learning a reward model from comparisons and updating a policy with PPO-style objectives under reference or KL control ~\citep{stiennon2020learning,schulman2017ppo,ouyang2022training}. DPO and related objectives remove the explicit RL loop by fitting likelihood-ratio preferences against a reference policy ~\citep{rafailov2024dpo}; RLAIF and group-relative variants alter the feedback source or variance-reduction scheme while retaining a policy-optimization view ~\citep{bai2022constitutional,shao2024deepseekmath}. In image generation, reward-gradient and preference methods adapt these ideas to diffusion or sampler structure: \DRaFT{} backpropagates differentiable rewards through a sampler ~\citep{clark2024draft}, diffusion-DPO adapts preference margins to diffusion likelihood bounds ~\citep{wallace2024diffusiondpo}, and ReNO or PAHI optimize sampling variables or prompt-specific noise distributions at test time ~\citep{eyring2024reno,kim2024pahi}. Recent preference-guidance methods also cast diffusion alignment as classifier-free-guidance-style inference ~\citep{jiang2026preferenceguidance}. These methods rely on a stochastic policy, a diffusion sampler, or test-time optimization to obtain the update direction. \DrPO{} instead estimates this direction directly in feature space from reward-ranked samples and reference samples.
\paragraph{Preference alignment.}
RLHF aligns generative policies by learning a reward model from comparisons and updating a policy with PPO-style objectives under reference or KL control~\citep{stiennon2020learning,schulman2017ppo,ouyang2022training}. DPO and related objectives remove the explicit RL loop by fitting likelihood-ratio preferences against a reference policy~\citep{rafailov2024dpo}; RLAIF and group-relative variants change the feedback source or normalization scheme while retaining a policy-optimization view~\citep{bai2022constitutional,shao2024deepseekmath}. In image generation, reward-gradient and preference methods adapt these ideas to diffusion or sampler structure: \DRaFT{} backpropagates differentiable rewards through a sampler~\citep{clark2024draft}, diffusion-DPO adapts preference margins to diffusion likelihood bounds~\citep{wallace2024diffusiondpo}, and ReNO or PAHI optimize sampling variables or prompt-specific noise distributions at test time~\citep{eyring2024reno,kim2024pahi}. Recent preference-guidance methods also cast diffusion alignment as classifier-free-guidance-style inference~\citep{jiang2026preferenceguidance}. \DrPO{} differs by estimating the update direction directly in feature space from reward-ranked samples and reference samples, without reward backpropagation, policy likelihoods, denoising trajectories, or test-time search.

\paragraph{Drifting models.}
% Drifting models introduce a likelihood-free training mechanism for one-step generators, where finite-batch drift in feature space replaces explicit density estimation or adversarial discrimination~\citep{deng2026drifting}. In their original formulation, the drift is built for distribution matching, using the local geometry between generated samples and samples from the data distribution. Our work brings this drifting view to reward-based post-training. The drift is no longer tied to unconditional data matching, but is instead defined by condition-specific preference information together with a frozen reference generator.
Drifting models~\citep{deng2026drifting} introduce a likelihood-free training mechanism for one-step generators, where finite-batch drift in feature space replaces explicit density estimation or adversarial discrimination. In their original formulation, the drift is used for distribution matching between generated samples and data samples. We use the same drifting view for reward-based post-training: the drift is defined by condition-specific preference information, while a frozen reference generator supplies the stability direction.

% \vspace{-2pt}
\section{Preliminaries}
% \vspace{-2pt}
\label{sec:prelim}

\subsection{Drifting Models}

Drifting models~\citep{deng2026drifting} train one-step generators by assigning a feature-space drift direction to each generated sample. For a condition $c$ and noise $\epsilon$, let $x=G_\theta(\epsilon,c)$ and let $f_\theta(\epsilon,c)$ be its feature representation. For a generated feature $x$, the drift moves the sample toward positive anchors and away from negative anchors:
\begin{equation}
\mathbf{V}_{\nu^+,\nu^-}(x)
=
\frac{1}{Z^+}\mathbb{E}_{b^+\sim\nu^+}\!\left[k(x,b^+)b^+\right]
-
\frac{1}{Z^-}\mathbb{E}_{b^-\sim\nu^-}\!\left[k(x,b^-)b^-\right],
\label{eq:prelim-drift-field}
\end{equation}
where $k$ measures feature similarity, $\nu^+$ denotes the positive distribution, $\nu^-$ denotes the negative distribution and $Z$'s are normalization constants. Given this field, the generator is trained against a detached drifted feature target:
\begin{equation}
\mathcal{L}_{\mathrm{drift}}(\theta)
=
\frac{1}{2}
\mathbb{E}_{\epsilon,c}
\left[
\left\|
f_\theta(\epsilon,c)
-
\operatorname{sg}\!\left(
f_\theta(\epsilon,c)+\mathbf{V}_{\nu^+,\nu^-}(f_\theta(\epsilon,c))
\right)
\right\|_2^2
\right],
\label{eq:main-prelim-loss}
\end{equation}
where $\operatorname{sg}(\cdot)$ denotes stop-gradient. Since the target is detached, this loss gives
\begin{equation}
\nabla_\theta \mathcal{L}_{\mathrm{drift}}
=
-
\mathbb{E}_{\epsilon,c}
\left[
\mathbf{V}_{\nu^+,\nu^-}(f_\theta(\epsilon,c))
\nabla_\theta f_\theta(\epsilon,c)
\right].
\label{eq:prelim-drift-policy-gradient}
\end{equation}
Thus the drifting loss can be read as a deterministic policy-gradient update whose feature-space velocity is synthesized from samples.

For minibatch training, Drifting models approximate the drift field with mean shift. Given a support batch $\mathcal{B}=\{b_i\}_{i=1}^{B}$, kernel weights are normalized within the batch:
\begin{equation}
w_i(x;\mathcal{B})
=
\frac{k(x,b_i)}{\sum_j k(x,b_j)},
\qquad
\widehat{\mu}_{\mathcal{B}}(x)
=
\sum_i w_i(x;\mathcal{B})b_i .
\end{equation}
With positive and negative batches $\mathcal{B}^+$ and $\mathcal{B}^-$, the approximate drift field is
\begin{equation}
\widehat{\mathbf V}(x)
=
\widehat{\mu}_{\mathcal{B}^+}(x)-\widehat{\mu}_{\mathcal{B}^-}(x).
\label{eq:main-prelim-drift}
\end{equation}
With an RBF kernel $k(x,b)=\exp(-s(x,b)/\tau)$, $w_i$ is the softmax weight over negative similarities.

% \begin{equation}
% \widehat{\mathbf V}(x)
% =
% \sum_i  \frac{k(x,b_i^+)}{\sum_j k(x,b_j^+)}b_i^+ - 
% \sum_i  \frac{k(x,b_i^-)}{\sum_j k(x,b_j^-)}b_i^-.
% \end{equation}

\subsection{Reinforcement Learning from Human Feedback}

Reinforcement Learning from Human Feedback (RLHF) aims to align the distribution with a preference dataset $\mathcal{D}=\{(c_i,x^+_i,x^-_i)\}_i$ where the positive sample $x^+_i$ is preferred to the negative sample $x^-_i$ under condition $c$. The standard procedure of RLHF trains a reward model $r_\psi(c,x)$ under the assumption of Bradley–Terry preference model~\citep{ouyang2022training}, with the following loss:
\begin{equation}
\mathcal{L}_{\mathrm{RM}}(\psi)
=
-
\mathop{\mathbb{E}}_{(c,x^+,x^-) \sim \mathcal{D}}
\log \sigma\!\left(
r_\psi(c,x^+) - r_\psi(c,x^-)
\right).
\end{equation}
With the reward model trained, the common alignment target is the tilted distribution that balances between the reward distribution and the pretrained distribution
\begin{equation}
q^*(x\mid c)
\propto
q_{\mathrm{ref}}(x\mid c)
\exp\!\left(r_\psi(c,x) / \lambda \right),
\label{eq:prelim-tilted}
\end{equation}
Fitting $q_\theta$ to this target is equivalent to the soft RL objective~\citep{haarnoja2018soft}
\begin{equation}
\max_{q_\theta}\;
\mathop{\mathbb{E}}_{c}
\mathop{\mathbb{E}}_{x\sim q_\theta(\cdot\mid c)}
\left[
r_\psi(c,x)
\right]
-
\lambda
\mathbb{E}_{c}
\left[
\mathrm{KL}\!\left(
q_\theta(\cdot\mid c)\,\|\,q_{\mathrm{ref}}(\cdot\mid c)
\right)
\right].
\label{eq:prelim-soft-rl}
\end{equation}
Notice that, for deterministic one-step generators, policy likelihoods and density ratios are not available in closed form. Performing principled RL training for one-step generators thus requires sample-based estimates.

% \vspace{-2pt}
\vspace{-1pt}
\section{Method}
\vspace{-1pt}
% \vspace{-2pt}
\label{sec:method}

% \begin{figure*}[t]
%   \centering
%   \begin{minipage}[t]{0.4\textwidth}
%     \footnotesize
%     \centering
%     \textbf{(a) Preference Velocity}\\[-0.2em]
%     \includegraphics[width=\linewidth]{figures/outputs/dual_drift_geometry_pref.pdf}
%   \end{minipage}
%   % \hfill
%   \begin{minipage}[t]{0.4\textwidth}
%     \footnotesize
%     \centering
%     \textbf{(b) Reference Drift Regularization}\\[-0.2em]
%     \includegraphics[width=\linewidth]{figures/outputs/dual_drift_geometry_ref.pdf}
%   \end{minipage}
%   \vspace{-4mm}
%   \caption{
%   \textbf{Drift construction in \DrPO{}.}
% (a) Reward-ranked positive and negative samples define the preference drift
% $V_{\mathrm{pref}}$, which moves generated features toward preferred samples
% and away from rejected samples. 
% (b) The final update direction combines this preference drift with the
% reference correction estimated from base and current samples,
% $V_{\mathrm{DrPO}}=V_{\mathrm{pref}}+\lambda V_{\mathrm{ref}}$.
%   }
%   \label{fig:dual-drift}
% \end{figure*}

\vspace{-2pt}
\subsection{Population-Rank Reward Model}
\vspace{-2pt}

% Let $x=g_\theta(\epsilon,c)$ and $z=\phi(x)$, where $\epsilon$ is the input noise,
% $x$ is the generated image, and $\phi$ is a fixed feature extractor. Suppose we
% are given positive and negative feature sets $\mathcal{A}^+$ and
% $\mathcal{A}^-$. We define a non-parametric reward model directly in feature space:
% \begin{equation}
% R_\text{dipole}(z)=\exp(E(z)),\qquad
% E(z)
% =
% \beta
% \sum_j
% \left[
% k(z,a_j^+)
% -
% k(z,a_j^-)
% \right],
% \label{eq:rank-reward}
% \end{equation}
% where $(a_j^+,a_j^-)$ is a preference pair, $k$ is a positive-definite kernel that decays with feature distance, and $\beta>0$ controls the field strength. 

% Differentiating Eqn.~\ref{eq:rank-reward} gives
% \begin{equation}
% \nabla_z\log R_\text{dipole}(z)
% =
% \beta\sum_j
% \left[
% k(z,a_j^+) \nabla_z\log k(z,a_j^+)
% -
% k(z,a_j^-) \nabla_z\log k(z,a_j^-)
% \right].
% \label{eq:rank-gradient}
% \end{equation}
% The induced field is kernel-weighted and local: points near positive samples are strongly attracted, points near negative samples are strongly repelled, and points far from the observed samples receive negligible update since $E(z)\approx 0$ and $R_\text{dipole}(z)\approx 1$. This locality is important for one-step finetuning because it lets the reward signal guide nearby samples while reducing extrapolated reward gradients far from the evidence in the current batch.

Let $x=g_\theta(\epsilon,c)$ and $z=\phi(x)$, where $\epsilon$ is the input noise, $x$ is the generated image, and $\phi$ is a fixed feature extractor. Given positive and negative feature samples $\mathcal{A}^+$ and $\mathcal{A}^-$, we define a non-parametric reward in feature space:
% \begin{equation}
% R_\text{dipole}(z)=\exp(E(z)),\qquad
% E(z)
% =
% \gamma
% \sum_{j=1}^{M}
% \left[
% k(z,a_j^+)
% -
% k(z,a_j^-)
% \right],
% \label{eq:rank-reward}
% \end{equation}
\begin{equation}
R_\text{rank}(z)=\exp(E(z)),\qquad
E(z)
=
\gamma
\left[
\log \sum_{j=1}^{M} k(z,a_j^+)
-
\log \sum_{j=1}^{M} k(z,a_j^-)
\right],
\label{eq:rank-reward}
\end{equation}
where $a_j^+\in\mathcal{A}^+$ and $a_j^-\in\mathcal{A}^-$ index equal-size positive and negative samples, $k$ is a feature-space kernel, and $\gamma>0$ sets the reward strength.

Intuitively, the two kernel sums aggregate evidence that samples near $z$ win or lose against randomly drawn opponents. Their ratio estimates the local win-to-loss odds. Under random pairing, the underlying odds increase monotonically with $z$’s percentile rank under the target reward, making $R_{\mathrm{rank}}$ a kernel-smoothed ordinal reward proxy.

Differentiating Eq.~\ref{eq:rank-reward} gives ($Z^+$, $Z^-$ are normalization constants):
% \begin{equation}
% \nabla_z\log R_\text{dipole}(z)
% =
% \gamma\sum_{j=1}^{M}
% \left[
% k(z,a_j^+) \nabla_z\log k(z,a_j^+)
% -
% k(z,a_j^-) \nabla_z\log k(z,a_j^-)
% \right].
% \label{eq:rank-gradient}
% \end{equation}
\begin{equation}
\nabla_z\log R_\text{rank}(z)
=
\frac{\gamma}{Z^+}\sum_{j=1}^{M}
\nabla_z k(z,a_j^+)
-
\frac{\gamma}{Z^-}\sum_{j=1}^{M}
\nabla_z k(z,a_j^-),
\label{eq:rank-gradient}
\end{equation}
% The induced field is kernel-weighted and local: points near positive samples are strongly attracted, points near negative samples are strongly repelled, and points far from the observed samples receive negligible update since when $E(z)\to 0$ we have $R_\text{dipole}(z) \to 1$. This locality is important for one-step finetuning because it lets the reward signal guide nearby samples while reducing extrapolated reward gradients far from the evidence in the current batch.
which, when an RBF kernel is used, is in the same form as drift fields.

\vspace{-2pt}
\subsection{Drifting Preference Optimization}
\vspace{-2pt}

For a differentiable, non-negative reward $R(z)$, let $x=g_\theta(\epsilon,c)$ and $z=\phi(x)$. The pathwise gradient of the reference-regularized soft RL objective is
\begin{equation}
\nabla_\theta J
=
\mathbb{E}_{\epsilon}
\left[
\left(
\nabla_z\log R(z)
+
\lambda
\left(
\nabla_z\log p_{\mathrm{base}}(z)
-
\nabla_z\log p_\theta(z)
\right)
\right)
\nabla_\theta z
\right],
\label{eq:drpo-policy-gradient}
\end{equation}
where $\nabla_\theta z=\nabla_x\phi(x)\nabla_\theta g_\theta(\epsilon,c)$.

Given an on-policy candidate batch $\mathcal{B}=\{x_i\}_{i=1}^{K}$ from $x_i=g_\theta(\epsilon_i,c)$, $\epsilon_i\sim\mathcal{N}(0,I)$, we evaluate the target reward and sample $M$ reward-ordered pairs $\mathcal{D}_\theta=\{(x_j^+,x_j^-)\}_{j=1}^{M}$ with $R(z_j^+)\geq R(z_j^-)$ and $z_j = \phi(x_j)$. The positive and negative samples are
\begin{equation}
\mathcal{A}^+=\{\phi(x_j^+)\}_{j=1}^{M},
\qquad
\mathcal{A}^-=\{\phi(x_j^-)\}_{j=1}^{M}.
\end{equation}
The reward-score term in Eq.~\ref{eq:drpo-policy-gradient} is replaced by the population-rank preference field (Eq.~\ref{eq:rank-reward}):
\begin{equation}
\nabla_z \log R(z)
\approx
V_{\mathrm{pref}}(z)
=
\nabla_z \log R_{\mathrm{rank}}(z;\mathcal{A}^+,\mathcal{A}^-).
\label{eq:drpo-rank-score}
\end{equation}

By treating samples for $\nabla_z \log p_\text{base}$ and $\nabla_z \log p_\theta(z)$ as positive and negative terms, respectively, we use the mean-shift estimate in Eq.~\ref{eq:main-prelim-drift} and write (with $\mathcal{R}$ and $\mathcal{Z}$ being independent sample sets from the two terms):
\begin{equation}
V_{\mathrm{ref}}(z_i)
=
\widehat{\mu}_{\mathcal{R}}(z_i)
-
\widehat{\mu}_{\mathcal{Z}}(z_i).
\label{eq:drpo-ref-drift}
\end{equation}
The final \DrPO{} field is
\begin{equation}
V_{\mathrm{DrPO}}(z_i)
=
V_{\mathrm{pref}}(z_i)
+
\lambda V_{\mathrm{ref}}(z_i).
\label{eq:drpo-total-drift}
\end{equation}

For $z_i=\phi(g_\theta(\epsilon_i,c))$, the detached target is
\begin{equation}
z_i^\star
=
\operatorname{sg}
\left(
z_i+\eta V_{\mathrm{DrPO}}(z_i)
\right),
\label{eq:drpo-target}
\end{equation}
with loss
\begin{equation}
\mathcal{L}_{\mathrm{DrPO}}(\theta)
=
\frac{1}{2K}
\sum_{i=1}^{K}
\left\|
z_i-z_i^\star
\right\|_2^2.
\label{eq:drpo-loss}
\end{equation}
The base model and feature extractor are used only during training. In practice, we adopt the same techniques in drifting models~\citep{deng2026drifting} to construct the drift field (details in Appendix~\ref{app:drift-implementation}).

\paragraph{Non-differentiable rewards.}
% \DrPO{} only requires the target reward $R$ to rank generated samples. Therefore,
% the same construction applies when $R$ is non-differentiable, such
% as rule-based evaluators or task-specific correctness checks: we use $R$ to form
% $\mathcal{A}^+$ and $\mathcal{A}^-$, and the subsequent dipole drift is computed
% entirely in feature space.
\DrPO{} only requires the target reward $R$ to rank generated samples. Therefore, the same construction applies when $R$ is non-differentiable, such as rule-based evaluators or task-specific correctness checks: we use $R$ to form $\mathcal{A}^+$ and $\mathcal{A}^-$, and the subsequent population-rank drift is computed entirely in feature space.

\paragraph{Offline variant.}
% The same update can be used with offline preference pairs. We replace the online
% feature sets $\mathcal{A}^+$ and $\mathcal{A}^-$ with chosen and rejected
% features from the dataset; the reference term and training loss remain
% unchanged. This keeps the drifting update identical, while removing the
% reward-ranking step used in the online setting.
We also make a naive attempt to extend \DrPO{} to offline preference finetuning. Given offline preference pairs, we replace the online feature sets $\mathcal{A}^+$ and $\mathcal{A}^-$ with chosen and rejected features from the dataset; the reference term and detached feature regression remain unchanged. This removes the online reward-ranking step while keeping the same drifting update.

\begin{algorithm}[h]
\caption{\DrPO{} (Online)}
\label{alg:drpo-impl}
\begin{algorithmic}[1]
\Require Generator $g_\theta$, frozen reference model $g_{\mathrm{ref}}$, feature map $\phi$, reward model $R$, drift radii $\Omega$, reference weight $\lambda$
\For{each optimization step}
  \State Sample condition $c$ and noises $\{\epsilon_i\}_{i=1}^{K}$
  \State Generate current samples $x_i=g_\theta(\epsilon_i,c)$ and reference samples $x_i^{\mathrm{ref}}=g_{\mathrm{ref}}(\epsilon_i,c)$
  \State Compute features $Z=\{\phi(x_i)\}_{i=1}^{K}$ and $\mathcal{R}=\{\phi(x_i^{\mathrm{ref}})\}_{i=1}^{K}$
  \State Rank $\{x_i\}_{i=1}^{K}$ with $R$ and form preference pairs $\mathcal{D}_\theta=\{(x_j^+,x_j^-)\}_{j=1}^{M}$
  \State Set $\mathcal{A}^+=\{\phi(x_j^+)\}_{j=1}^{M}$ and $\mathcal{A}^-=\{\phi(x_j^-)\}_{j=1}^{M}$
  \State Construct drift field: $V_{\mathrm{pref}}\gets\Call{DriftField}{Z,\mathcal{A}^+,\mathcal{A}^-,\Omega}$
  \State Construct drift field: $V_{\mathrm{ref}}\gets\Call{DriftField}{Z,\mathcal{R},\operatorname{sg}(Z),\Omega}$
  \State $V_{\mathrm{DrPO}}\gets V_{\mathrm{pref}}+\lambda V_{\mathrm{ref}}$
  \State $Z^\star\gets\operatorname{sg}\!\left(Z+\eta V_{\mathrm{DrPO}}\right)$
  \State Update $\theta$ with $\mathcal{L}_{\mathrm{DrPO}}=\frac{1}{K}\sum_i\|z_i-z_i^\star\|_2^2$
\EndFor
\end{algorithmic}
\end{algorithm}

% \vspace{-3pt}
\vspace{-1pt}
\section{Experiments}
% \vspace{-2pt}
\vspace{-1pt}
\label{sec:exp}
\label{sec:experimental-setup}

\subsection{Experimental Setup}
\label{sec:exp-setup}

\noindent\textbf{Base models.}
We use SD-Turbo~\citep{sauer2024add} (\textit{license: Stability AI Community License}) as the main one-step text-to-image base model, and evaluate SDXL-Turbo~\citep{podell2023sdxl,sauer2024add} (\textit{license: Stability AI Community License}) in the base-model ablation. The two models use different architectures and latent spaces, so the SDXL-Turbo experiment tests whether the same population-rank-drift update transfers beyond the SD-Turbo setting.

\noindent\textbf{Target reward models.}
The default target reward is PickScore~\citep{kirstain2024pickapic} (\textit{license: MIT}). We also use HPSv3~\citep{ma2025hpsv3} (\textit{license: MIT}) to test training with a large VLM-based reward. In the target-reward ablation, we replace or combine PickScore with Aesthetic Score (AES)~\citep{laionaes} (\textit{license: MIT}), HPSv2~\citep{wu2023hpsv2} (\textit{license: Apache-2.0}), ImageReward (IR)~\citep{xu2024imagereward} (\textit{license: Apache-2.0}), and CLIP~\citep{radford2021clip} (\textit{license: MIT}). AES is unconditional, while the other rewards are text-conditioned.

\noindent\textbf{Prompts and evaluation sets.}
Online experiments use prompts from the Pick-a-Pic v2 training split~\citep{kirstain2024pickapic}. The primary evaluation uses the Pick-a-Pic v2 test split with $424$ prompts and five matched seeds. We also evaluate on Parti-Prompts~\citep{yu2022parti} ($1{,}632$ prompts included).

\noindent\textbf{GenEval sets.}
For non-differentiable reward experiments, we use $553$ GenEval prompts covering object presence, counting, color, position, and attribute binding~\citep{ghosh2023geneval}. For the GenEval training set, we train separately on each subtask and evaluate on the corresponding test prompts  independently. Training prompts are generated from the corresponding task templates. All training prompts present in the test set are removed.

\noindent\textbf{Baselines.}
We compare with two groups of finetuning methods: (1) \emph{reward-gradient-based methods}, including \DRaFT{} and \VGGFlow{}~\citep{clark2024draft,liu2025value}; and (2) \emph{reward-gradient-free methods}, including \DPOone{}, \PSO{}, and \GRPOone{}~\citep{rafailov2024dpo,wallace2024diffusiondpo,shao2024deepseekmath}. For methods not originally designed for deterministic one-step generators, we adapt their core objectives to the same finetuning setting and evaluate with matched prompts and seeds. Details of the one-step adaptations are given in the appendix.

\noindent\textbf{Metrics.}
We evaluate text-to-image generation with scalar preference and quality metrics, including PickScore, AES, HPSv2, HPSv3, and ImageReward. FID is reported when matched references are available~\citep{heusel2017fid}; GenEval uses the standard compositional scores~\citep{ghosh2023geneval}. We also report pairwise win rates using Qwen3-VL~\citep{qwen2025qwen3vl} as an external VLM judge, following recent VQA/MLLM-based evaluation protocols~\citep{hu2023tifa,lin2024vqascore,ku2024viescore,lee2024prometheusvision}. Qwen3-VL compares image pairs generated from the same prompt and outputs \texttt{A}, \texttt{B}, or \texttt{tie}; we randomize order to reduce position bias~\citep{zheng2023judging,shi2025systematic} and aggregate parsed JSON decisions into win/loss/tie rates. Details are provided in Appendix~\ref{app:qwen3vl-prompt}.

\noindent\textbf{Feature extractors.}
Our default feature extractor is the 340M-parameter latent-MAE encoder from Drifting models~\citep{deng2026drifting}. We ablate this choice against raw latent features and DINOv2 features extracted from VAE-decoded images. We also compare multi-layer latent-MAE variants: latent-MAE$_1$ is the default multi-feature variant, latent-MAE$_2$ uses the last three layers, and latent-MAE$_3$ uses only the final layer. For SDXL-Turbo, where the SD-VAE-specific latent-MAE is not directly applicable, we use decoded image-space MAE features~\citep{he2022masked}.

\noindent\textbf{Construction of the drift field.} Following \citet{deng2026drifting}, we estimate the drift field from mini-batches by combining attraction toward positive samples and repulsion from negative samples across multiple neighborhood scales. We adopt their normalization and stabilization techniques for improved optimization. Implementation details are provided in Appendix~\ref{app:drift-implementation}.

\noindent\textbf{Implementation details.}
We finetune LoRA parameters~\citep{hu2022lora} with rank $16$. Unless otherwise specified, training uses a per-GPU mini-batch size of $24$ on $4$ GPUs with $8$ gradient-accumulation steps, giving an effective batch size of $768$. Each \DrPO{} mini-batch uses one prompt to generate $24$ candidates; the reward model samples $12$ reward-ordered pairs to construct $\mathcal{A}^+$ and $\mathcal{A}^-$. We use AdamW with learning rate $1\times10^{-4}$. For HPSv3 experiments, we use learning rate $3\times10^{-5}$ and reduce the number of candidate images per prompt to $16$. For the \DRaFT{}+HPSv3 baseline, we use mini-batch size $2$ with $64$ gradient-accumulation steps to avoid out-of-memory errors while keeping the same effective batch size.

\vspace{-2pt}

\vspace{-2pt}
\subsection{Results}
\vspace{-2pt}
% \vspace{-2pt}
\label{sec:results}

\begin{table}[t]
  \centering
%   \caption{\textbf{Results on SDXL-Turbo.}
% We report test-time generator calls as inference steps and indicate whether each method backpropagates through the target reward during training. PS denotes PickScore $\times 100$ and IR denotes ImageReward $\times 10$.
% }
  \setlength{\tabcolsep}{3.2pt}
  \renewcommand{\arraystretch}{1.0}
  \footnotesize
  % \resizebox{0.86\textwidth}{!}{%
  \begin{tabular}{@{}l c c ccc ccc@{}}
    \toprule
    \multirow{2}{*}{Method} &
    \multirow{2}{*}{\makecell[c]{Inference\\Steps}} &
    \multirow{2}{*}{\makecell[c]{Reward\\Grad}} &
    \multicolumn{3}{c}{Pick-a-Pic v2 Test} &
    \multicolumn{3}{c}{Parti-Prompts} \\
    \cmidrule(lr){4-6}
    \cmidrule(lr){7-9}
    & & &
    PS $\uparrow$ &
    AES $\uparrow$ &
    IR $\uparrow$ &
    PS $\uparrow$ &
    AES $\uparrow$ &
    IR $\uparrow$ \\
    \midrule
SDXL base & 50 & -- & 22.15 & 6.104 & 6.85 & 22.64 & 5.761 & 7.24 \\
SDXL-DPO  & 50  & -- & 22.57 & 6.076 & 9.38 & 22.95 & 5.811 & 10.66 \\

    SDXL-Turbo (base) & 1 & -- & 22.45 & 6.059 & 9.36 & 22.77 & 5.693 & 9.13 \\

    \midrule

    \DRaFT{}   & 1 & Yes & \textbf{24.45} & \underline{6.712} & \textbf{12.70} & \textbf{24.34} & \underline{6.485} & \textbf{12.66} \\
    \VGGFlow{} & 1 & Yes & \underline{24.27} & 6.490 & 12.19 & \underline{23.98} & 6.200 & 11.40 \\

    \midrule

    \DPOone{}  & 1 & No & 22.77 & 6.227 & 10.44 & 22.85 & 6.019 & 11.25 \\
    \PSO{} & 1 & No & 22.56 & 6.092 & 8.97 & 22.87 & 5.744 & 9.17 \\
    \GRPOone{} & 1 & No & 22.50 & 6.077 & 9.57 & 22.80 & 5.710 & 9.27 \\
    \rowcolor{gray!12}
    \textbf{\DrPO{}} & 1 & No & 23.66 & \textbf{6.717} & \underline{12.46} & 23.71 & \textbf{6.665} & \underline{12.60} \\
    \bottomrule
  \end{tabular}
  % }
    \vspace{1mm}
    \caption{
        \textbf{Results on SDXL-Turbo.}
        We report test-time generator calls as inference steps and indicate whether each method backpropagates through the target reward during training. For multi-step SDXL baselines, $50\times2$ denotes classifier-free guidance with two model evaluations per denoising step. PS denotes PickScore $\times 100$ and IR denotes ImageReward $\times 10$.
    }
  \label{tab:sdxl-turbo-main-results}
  \vspace{-5mm}
\end{table}

\begin{figure}[t]
\centering
\begin{minipage}[t]{0.49\linewidth}
    \centering
    \includegraphics[width=\linewidth]{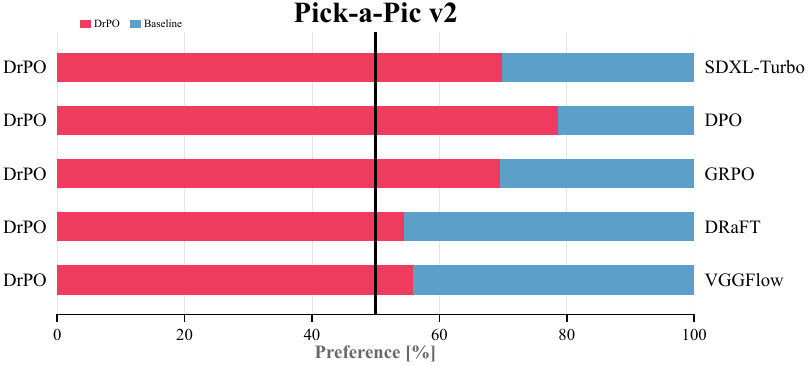}
    % \caption*{Pick-a-Pic v2}
\end{minipage}
\hfill
\begin{minipage}[t]{0.49\linewidth}
    \centering
    \includegraphics[width=\linewidth]{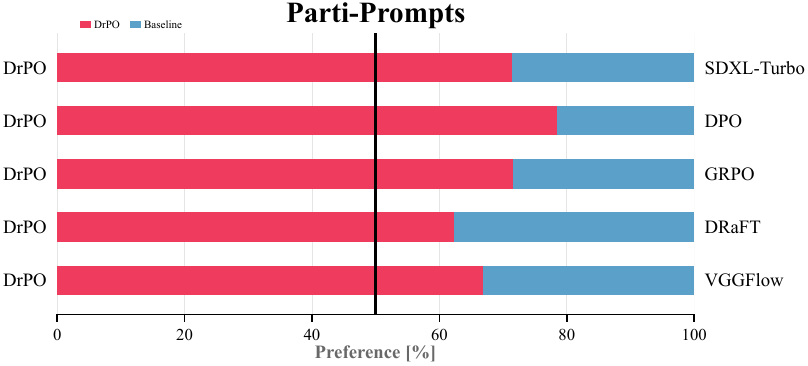}
    % \caption*{Parti-Prompts}
\end{minipage}
\caption{ Qwen3-VL pairwise preference evaluation. For each prompt, Qwen3-VL compares two matched generations under the same prompt and selects the cleaner, more coherent prompt realization; ties are allowed and A/B order is randomized. Red indicates \DrPO{} preference, blue indicates the compared baseline preference, and the vertical line marks 50\%. The full judge prompt is provided in Appendix~\ref{app:qwen3vl-prompt}.}
\label{fig:qwen3vl_drpo_preference_by_dataset}
% \vspace{-12mm}
\vspace{-2mm}
\end{figure}

\begin{table}[t]
  \centering
  \setlength{\tabcolsep}{3.2pt}
  \renewcommand{\arraystretch}{1.0}
  \footnotesize
  % \resizebox{0.86\textwidth}{!}{%
  \begin{tabular}{@{}l c c ccc ccc@{}}
    \toprule
    \multirow{2}{*}{Method} &
    \multirow{2}{*}{\makecell[c]{Inference\\Steps}} &
    \multirow{2}{*}{\makecell[c]{Reward\\Grad}} &
    \multicolumn{3}{c}{Pick-a-Pic v2 Test} &
    \multicolumn{3}{c}{Parti-Prompts} \\
    \cmidrule(lr){4-6}
    \cmidrule(lr){7-9}
    & & &
    PS $\uparrow$ &
    AES $\uparrow$ &
    IR $\uparrow$ &
    PS $\uparrow$ &
    AES $\uparrow$ &
    IR $\uparrow$ \\
    \midrule

    SD1.5      & 50 & --  & 20.79 & 5.455 &  1.22 & 21.49 & 5.358 &  2.25 \\
    LCM-SD1.5  & 4  & --  & 20.50 & 5.497 & -3.08 & 21.15 & 5.396 & -1.94 \\
    SD2.1      & 50 & --  & 21.09 & 5.645 &  2.49 & 21.77 & 5.547 &  3.97 \\
    SD-Turbo (base)         & 1  & --  & 21.88 & 6.054 &  5.75 & 22.29 & 5.758 &  5.37 \\

    \midrule

    \DRaFT{}             & 1  & Yes & \textbf{24.69} & \textbf{6.820} &  \underline{9.63} & \textbf{23.07} & \textbf{6.516} &  \underline{7.72} \\
    \VGGFlow{}           & 1  & Yes & \underline{23.73} & 6.378 &  7.74 & \underline{22.99} & 6.027 &  6.50 \\

    \midrule

    \DPOone{}            & 1  & No  & 22.02 & 6.080 &  5.95 & 22.39 & 5.793 &  5.21 \\
    \PSO{}               & 1  & No  & 21.88 & 6.059 &  5.80 & 22.29 & 5.763 &  5.42 \\
    \GRPOone{}           & 1  & No  & 21.98 & 6.077 &  6.08 & 22.35 & 5.779 &  5.65 \\
    % \rowcolor{gray!12}\textbf{\DrPO{} w/o ref.}     & 1  & No  & \underline{23.55} & \underline{6.603} & \textbf{10.07} & \underline{23.02} & \underline{6.384} &  \textbf{7.89} \\
    \rowcolor{gray!12}
    \textbf{\DrPO{}}
                          & 1  & No  & 23.49 & \underline{6.485} &  \underline{9.54} & \underline{22.99} & \underline{6.284} &  \underline{7.46} \\
    \bottomrule
  \end{tabular}
  % }
  \vspace{1mm}
  \caption{\textbf{Results on SD-Turbo.}
    % We report test-time generator calls as inference steps and indicate whether each method backpropagates through the target reward during training. Daggered baselines use official Diffusers recipes.
    We report test-time generator calls as inference steps and indicate whether each method backpropagates through the target reward during training. PS denotes PickScore $\times 100$ and IR denotes ImageReward $\times 10$.
}
  \label{tab:main-results-2}
  % \vspace{-10pt}
  \vspace{-12pt}
  % \vspace{5mm}
\end{table}

\begin{figure}[t]
  \centering
  \vspace{-1mm}
  \includegraphics[width=0.95\textwidth]{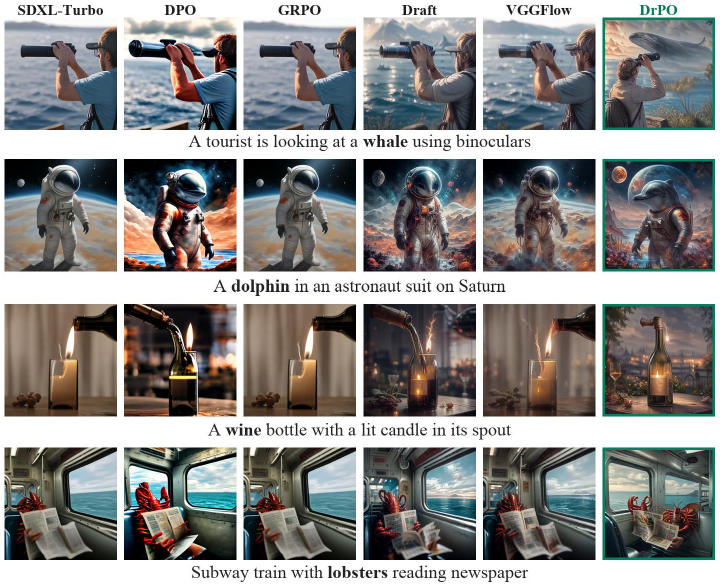}
  \caption{Qualitative comparison on Pick-a-Pic v2 prompts for SDXL-Turbo. Rows compare the SDXL-Turbo base, one-step preference baselines, and \DrPO{} variants under matched prompts and seeds. The examples complement the SDXL-Turbo quantitative comparison in Table~\ref{tab:sdxl-turbo-main-results}.}
  % \vspace{-13mm}
  \label{fig:sdxl-qualitative-comparison}
\end{figure}

\begin{figure}[t]
  \centering
  \includegraphics[width=0.95\textwidth]{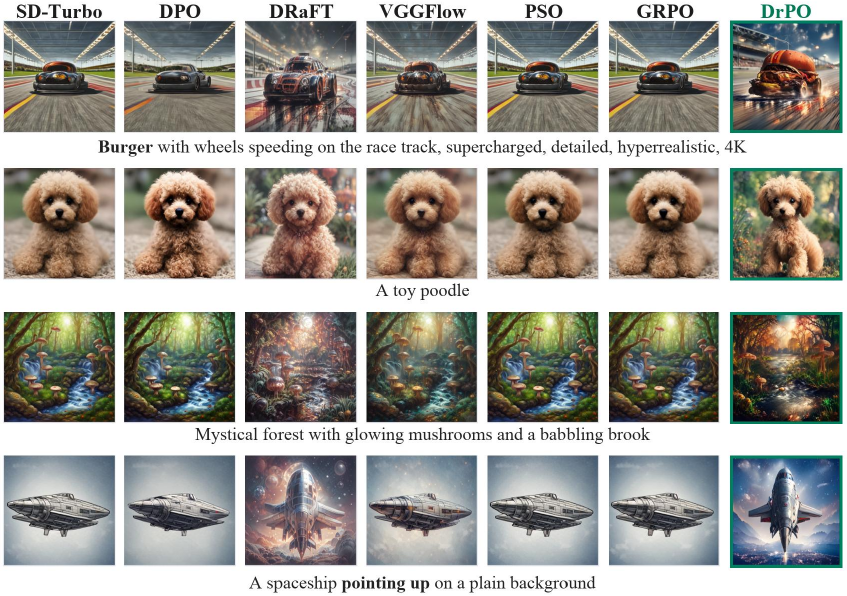}
  \caption{Qualitative headline comparison on selected Pick-a-Pic v2 prompts for SD-Turbo. The grid follows the same row-caption layout as the SDXL-Turbo qualitative comparison. \DrPO{} improves preference-aligned visual quality over the SD-Turbo base while preserving one-step inference.}
  \label{fig:qualitative-comparison}
  \vspace{-2mm}
\end{figure}

\leavevmode
\begin{wrapfigure}[7]{r}{0.32\textwidth}
  \vspace{-42pt}
  \centering
  \includegraphics[width=0.95\linewidth]%
    {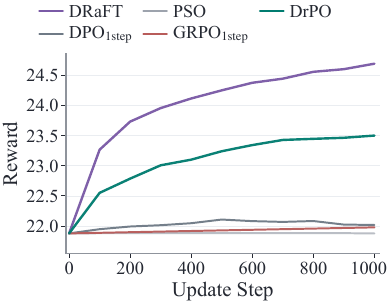}
  \vspace{-4pt}
  \caption{Reward curves on SD-Turbo.}
  \label{fig:reward-conv}
  \vspace{-5pt}
\end{wrapfigure}
% \leavevmode
% \begin{wrapfigure}{r}{0.35\textwidth}
%   % \vspace{-30pt}
%   \vspace{-25pt}
%   \centering
%   % \includegraphics[width=\linewidth]{figures/table1_update_step_reward_curves.pdf}
%   \includegraphics[width=0.95\linewidth]{figures/outputs/Reward_convergence_curves.pdf}
%   \vspace{-2pt}
%   \caption{Reward curves on SD-Turbo.}
%   \label{fig:reward-conv}
%   \vspace{-50pt}
% \end{wrapfigure}
\noindent\textbf{Main results.} 
Table~\ref{tab:sdxl-turbo-main-results} reports results on SDXL-Turbo, an SDXL-based one-step generator. Using the target reward only for ranking, \DrPO{} consistently improves PickScore, AES, and ImageReward on both Pick-a-Pic v2 and Parti-Prompts, with the strongest gains among reward-gradient-free baselines. Although direct reward-gradient methods such as \DRaFT{} achieve higher scores in some columns, they tend to produce visible artifacts (Fig.~\ref{fig:qualitative-comparison}) and perform worse under Qwen3-VL evaluation (Sec.~\ref{sec:qwen3vl-results}). Table~\ref{tab:main-results-2} demonstrates that these gains transfer to other models. Qualitative comparisons are provided in Figs.~\ref{fig:sdxl-qualitative-comparison}, \ref{fig:qualitative-comparison}, and \ref{fig:qualitative-comparison-full}.

% \clearpage

\noindent\textbf{Qwen3-VL pairwise preference.} 
\label{sec:qwen3vl-results}
% Figure~\ref{fig:qwen3vl_drpo_preference_by_dataset} complements scalar rewards with a VLM-based pairwise judge comparing two images from the same prompt. The judge prioritizes semantic fidelity, global coherence, artifact avoidance, and aesthetics, and outputs a structured decision with ties allowed. Unlike PickScore or ImageReward, this evaluation measures whether an image is a clean and coherent realization of the prompt. The red bars show that \DrPO{} is consistently preferred over one-step baselines on both Pick-a-Pic v2 and Parti-Prompts, indicating stronger prompt realization under an external VLM evaluator. The judge prompt is provided in Appendix~\ref{app:qwen3vl-prompt}.
Figure~\ref{fig:qwen3vl_drpo_preference_by_dataset} reports pairwise judgments from Qwen3-VL for matched generations under the same prompt. Unlike scalar reward metrics, this evaluation asks a VLM judge to compare two complete images with respect to prompt fidelity, visual coherence, artifacts, and aesthetics. The judge outputs \texttt{A}, \texttt{B}, or \texttt{tie}; A/B order is randomized, and decisions are aggregated into win/loss/tie rates. The red bars show that \DrPO{} is consistently preferred over one-step baselines on both Pick-a-Pic v2 and Parti-Prompts, indicating better prompt realization under an external VLM evaluator. The judge prompt is provided in Appendix~\ref{app:qwen3vl-prompt}.

% \noindent\textbf{SD-Turbo comparison.} Table~\ref{tab:main-results} compares \DrPO{} with reward-backpropagation methods and one-step preference baselines on SD-Turbo. Reward-backpropagation methods directly optimize the scalar reward and therefore obtain the strongest reward numbers in several columns, but they require differentiating through the target reward. \DrPO{} uses the target reward only to form positive and negative supports and gives the largest improvement over SD-Turbo among reward-gradient-free one-step preference methods. Matched qualitative examples are shown in Figures~\ref{fig:qualitative-comparison} and~\ref{fig:reward-selector-matrix}, with additional grids in Appendix Figures~\ref{fig:qualitative-comparison-full} and \ref{fig:reward-model-matrix-full}.

\begin{wrapfigure}{r}{0.57\columnwidth}
  \vspace{-10pt}
  \centering
  \begin{minipage}[t]{0.48\linewidth}
    \vspace{0pt}
    \centering
    \tiny
    \setlength{\tabcolsep}{2.4pt}
    \renewcommand{\arraystretch}{1.12}
    \resizebox{\linewidth}{!}{%
    \begin{tabular}{@{}lcc@{}}
      \toprule
      Metric &
      \makecell[c]{\DRaFT{}} &
      \makecell[c]{\textbf{\DrPO{}}} \\
      \midrule
      Reward grad & Yes & No \\
      \makecell[l]{Effective batch} & 192 & 192 \\
      Update time (s) $\downarrow$ & 21.62 & \textbf{6.17} \\
      Speedup $\uparrow$ & 1.00$\times$ & \textbf{3.51$\times$} \\
      Backward (s) & 9.99 & \textbf{0.34} \\
      \bottomrule
    \end{tabular}
    }
  \end{minipage}
  \hfill
  \begin{minipage}[t]{0.48\linewidth}
    \vspace{0pt}
    \centering
    \includegraphics[width=\linewidth]{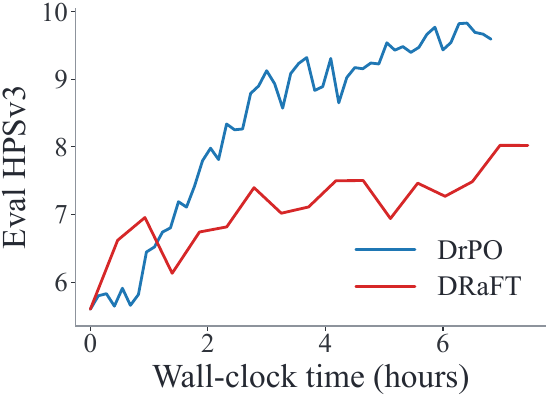}
  \end{minipage}
  \vspace{-0.2em}
  \caption{Efficiency comparison on HPSv3.}
  \label{fig:hpsv3-speed-profile}
  % \vspace{-14pt}
  \vspace{-18pt}
\end{wrapfigure}

\noindent\textbf{Efficiency with large reward models.}
% Figure~\ref{fig:hpsv3-speed-profile} profiles training with HPSv3. Since \DrPO{} uses HPSv3 for ranking rather than backpropagation, the reward-model backward pass is removed from each update, resulting in $3.51\times$ speedup.
Figure~\ref{fig:hpsv3-speed-profile} profiles training with HPSv3. Under the matched effective-batch setting, \DrPO{} removes HPSv3 reward-model backpropagation and reduces update time from $21.62$s to $6.17$s, giving a $3.51\times$ speedup.

\begin{figure}[t]
  \vspace{-4pt}
  \centering

  \begin{minipage}[t]{0.47\linewidth}
    \vspace{3pt}
    \centering
    \captionsetup{type=table}

    \small
    \setlength{\tabcolsep}{6pt}
    \renewcommand{\arraystretch}{1.12}

    \begin{tabular}{@{}lcc@{}}
      \toprule
      Metric & SD-Turbo & \shortstack{\textbf{+\DrPO{}}} \\
      \midrule
      Single $\uparrow$      & 98.8 & \textbf{100.0} \\
      Two $\uparrow$         & 46.5 & \textbf{55.6} \\
      Count $\uparrow$       & 33.8 & \textbf{42.5} \\
      Colors $\uparrow$      & 83.8 & \textbf{87.2} \\
      Position $\uparrow$    & 8.0  & \textbf{13.0} \\
      Color Attr. $\uparrow$ & 9.0  & \textbf{13.0} \\
      \bottomrule
    \end{tabular}
    \caption{Performance of \DrPO{} on GenEval tasks.}
    \label{tab:online-drpo-geneval}
  \end{minipage}
  \hfill
  \begin{minipage}[t]{0.50\linewidth}
    \vspace{0pt}
    \centering
    \includegraphics[width=\linewidth]{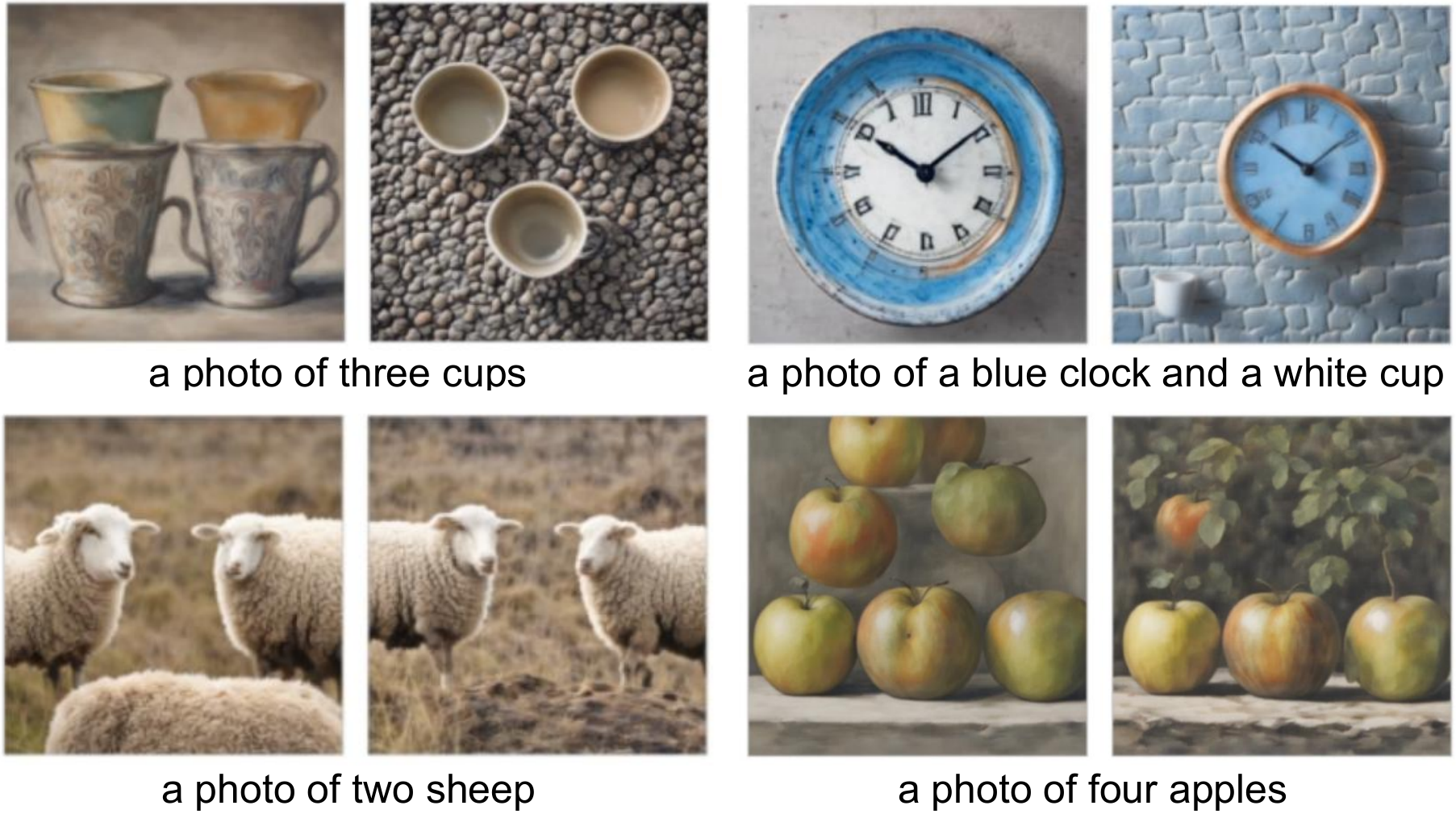}

    \caption{Generated images for \DrPO{} on GenEval prompts.}
    \label{fig:online-drpo-geneval}
  \end{minipage}

  % \vspace{-2pt}
\end{figure}

\noindent\textbf{Non-differentiable rewards.}
% Figure~\ref{fig:online-drpo-geneval} evaluates \DrPO{} with GenEval scores as the target reward. GenEval provides non-differentiable task-level scores for compositional constraints such as counting, color, position, and attribute binding~\citep{ghosh2023geneval}. Because \DrPO{} only requires the target reward to rank candidates, the same preference pair construction applies without modification.
Figure~\ref{fig:online-drpo-geneval} evaluates \DrPO{} with GenEval scores as the target reward. GenEval provides non-differentiable task-level scores for compositional constraints such as counting, color, position, and attribute binding~\citep{ghosh2023geneval}. \DrPO{} uses these scores to rank candidates and form $\mathcal{A}^+$ and $\mathcal{A}^-$, with no change to the feature-space drift update.

% \newpage

\begin{figure}[t]
  \centering
  \includegraphics[width=0.95\textwidth]{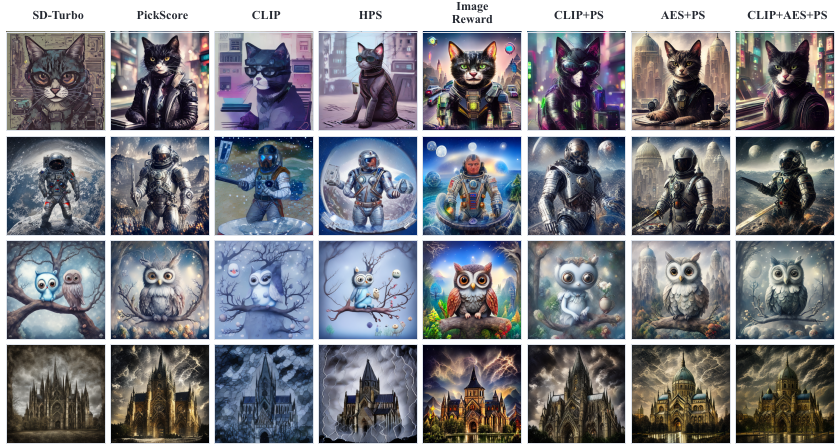}
  \caption{Qualitative comparison across reward models. Each row generated with identical prompt; metrics in Table~\ref{tab:reward-generality}.}
  \label{fig:reward-selector-matrix}
  % \vspace{-8pt}
  \vspace{-10pt}
\end{figure}

\noindent\textbf{Ablation on target rewards.}
% Figure~\ref{fig:reward-selector-matrix} and Table~\ref{tab:reward-generality} replace the target reward used to form positive and negative samples. The optimized behavior follows the chosen target: PickScore gives the largest PickScore gains, HPS rewards improve HPS-based metrics, and AES or ImageReward shifts the model toward their respective scores. The drift update is unchanged across these ablations.
% Figure~\ref{fig:reward-selector-matrix} and Table~\ref{tab:reward-generality} replace the target reward used to form $\mathcal{A}^+$ and $\mathcal{A}^-$. 
% The drift estimator and optimization objective are otherwise kept fixed. The results show that DrPO can translate different reward signals into effective  updates, indicating that is not tied to a specific reward model but provides a general interface for preference-driven one-step generator finetuning.
% The optimized behavior follows the chosen reward: PickScore gives the largest PickScore gains, HPS rewards improve HPS-based metrics, and AES or ImageReward shifts the model toward their respective scores. The drift update is unchanged across these runs.
Figure~\ref{fig:reward-selector-matrix} and Table~\ref{tab:reward-generality} show \DrPO{} results with different target reward models. The drift estimator and optimization objective are kept fixed, suggesting that \DrPO{} is robust to the choice of reward model.

\noindent\textbf{Ablation on drift design.}
% Table~\ref{tab:ablation} shows the ablation results on candidate count, feature extractor, kernel choice and the value of $\beta$. Increasing the candidate count improves PickScore and AES, suggesting that a larger on-policy pool provides more reliable gradient estimates. For feature extractors, latent-MAE features perform better than raw latent features, while using raw latents substantially degrades performance; this indicates that the drift benefits from a representation where local neighborhoods capture semantic image changes. We observe that \DrPO{} is not highly sensitive to the choice of kernels. On the choice of $\beta$, we find that increasing $\beta$ does not necessarily lead to better performance, possibly due to the relatively hard optimization process compared to other methods.
Table~\ref{tab:ablation} ablates candidate count, feature extractor, kernel choice, and velocity scale. Increasing the candidate count improves PickScore and AES, suggesting that a larger on-policy pool gives more stable pairwise estimates. Latent-MAE features outperform raw latent features, while raw latents substantially degrade performance; the drift benefits from a representation whose local neighborhoods track semantic image changes. While RBF kernel is the only theoretically reasonable choice for \DrPO{}, using other kernels may empirically be better. For the velocity scale $\eta$, larger values do not necessarily improve performance.

\begin{table}[t]
  \centering
  % \scriptsize
  % \setlength{\tabcolsep}{1.8pt}
  \renewcommand{\arraystretch}{0.84}
  \begin{minipage}[t]{0.23\textwidth}
  \centering
  \resizebox{\linewidth}{!}{%
  \renewcommand{\arraystretch}{1.29}
  \begin{tabular}{@{}lcc@{}}
    \toprule
    Candidates & PS $\uparrow$ & AES $\uparrow$ \\
    \midrule
    \rowcolor{gray!15} SD-Turbo & 21.88 & 6.054  \\
    $K=16$ & 23.24 & 6.409 \\
    $K=24$ & 23.53 & 6.552  \\
    $K=32$ & \textbf{23.57} & \textbf{6.599}  \\
    \bottomrule
  \end{tabular}
  }
  \par\vspace{2pt}
  \parbox{\linewidth}{\centering\scriptsize {(a) Batch generation}}
  \end{minipage}
  \hfill
  \begin{minipage}[t]{0.23\textwidth}
  \centering
  \renewcommand{\arraystretch}{1.19}
  \resizebox{\linewidth}{!}{%
  \begin{tabular}{@{}lcc@{}}
    \toprule
    Feature & PS $\uparrow$ & AES $\uparrow$  \\
    \midrule
    \rowcolor{gray!15} SD-Turbo & 21.88 & 6.054 \\
    latent-MAE$_1$ & \textbf{23.55} & 6.513 \\
    latent-MAE$_2$ & 23.50 & \textbf{6.526}  \\
    latent-MAE$_3$ & 23.48 & 6.506 \\
    Latent & 20.52 & 4.543 \\
    VAE-dec.+DINOv2 & 22.28 & 6.252 \\
    \bottomrule
  \end{tabular}
  }
  \par\vspace{2pt}
  \parbox{\linewidth}{\centering\scriptsize {(b) Feature map}}
  \end{minipage}
  \hfill
  \begin{minipage}[t]{0.23\textwidth}
  \centering
  \resizebox{\linewidth}{!}{%
  \renewcommand{\arraystretch}{1.10}
  \begin{tabular}{@{}lcc@{}}
    \toprule
    Kernel & PS $\uparrow$ & AES $\uparrow$ \\
    \midrule
    \rowcolor{gray!15} SD-Turbo & 21.88 & 6.054\\
    Cosine & \textbf{23.63} & 6.509  \\
    RBF & 23.50 & 6.590\\
    Exponential & 23.53 & \textbf{6.594}  \\
    Laplacian & 23.51 & 6.515  \\
    \bottomrule
  \end{tabular}
  }
  \par\vspace{2pt}
  \parbox{\linewidth}{\centering\scriptsize {(c) Kernel}}
  \end{minipage}
  \hfill
  \begin{minipage}[t]{0.23\textwidth}
  \centering
  \renewcommand{\arraystretch}{1.16}
  \resizebox{\linewidth}{!}{%
  
  \begin{tabular}{lrrr}
    \toprule
    Weight & PS $\uparrow$ & AES $\uparrow$ \\
    \midrule
    \rowcolor{gray!15} SD-Turbo & 21.88 & 6.054 \\
    $\beta=1000$ & 23.51 & 6.510 \\
    $\beta=3000$ & \textbf{23.53} & \textbf{6.542}  \\
    $\beta=5000$ & 23.51 & 6.485  \\
    $\beta=10000$ & 23.46 & 6.444  \\
    \bottomrule
  \end{tabular}
  }
  \par\vspace{2pt}
  \parbox{\linewidth}{\centering\scriptsize {(d) Velocity scale $\eta$}}
  \end{minipage}

  \vspace{2mm}
  \label{tab:main-results}
  \caption{Ablations on different design choices.}
  \label{tab:ablation}
  \vspace{-7mm}
\end{table}

\begin{wrapfigure}{r}{0.33\textwidth}
  % \vspace{-50pt}
  \vspace{-25pt}
  \centering
  \includegraphics[width=0.8\linewidth]{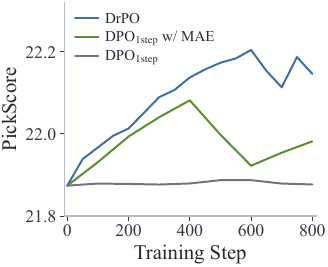}
  \vspace{-0.5em}
  \caption{Offline \DrPO{} Convergence.}
  \label{fig:offline-drpo-dpo-pickscore}
  % \vspace{-0.9em}
  \vspace{-30pt}
\end{wrapfigure}

\noindent\textbf{Ablation on reference regularization.}
% Table~\ref{tab:reference-final} compares two reference terms. Although related to perceptual regularization, the proposed reference drift plays a different role: perceptual loss anchors each generated sample to its paired reference image, whereas reference drift constructs $V_{\mathrm{ref}}$ from local supports of the base and current generators and combines it with the preference drift. Reference drift yields small but consistent gains over the perceptual baseline on most alignment metrics, suggesting that preserving the reference distribution is more effective than pairwise feature anchoring.
Table~\ref{tab:reference-final} compares the proposed reference drift with a perceptual regularization baseline. Perceptual loss anchors each generated sample to its paired reference image, while reference drift uses local neighborhoods from the base and current generators in the same feature geometry. Reference drift gives small but consistent gains over the perceptual baseline on most alignment metrics.

% \begin{wrapfigure}{r}{0.35\textwidth}
%   % \vspace{-50pt}
%   \centering
%   \includegraphics[width=\linewidth]{figures/outputs/offline_drpo_dpo_pickscore_curve.pdf}
%   \vspace{-1em}
%   \caption{Offline \DrPO{} Convergence.}
%   \label{fig:offline-drpo-dpo-pickscore}
%   % \vspace{-0.9em}
%   \vspace{-18pt}
% \end{wrapfigure}

\begin{center}
  \begin{minipage}[h]{0.53\textwidth}
    \centering
    \vspace{0pt}
    \scriptsize
    \setlength{\tabcolsep}{2.0pt}
    \renewcommand{\arraystretch}{0.98}

    \begin{tabular}{lcccc}
      \toprule
      Objective &
      \makecell{PickScore $\uparrow$} &
      \makecell{CLIP $\uparrow$} &
      AES $\uparrow$ &
      \makecell{HPSv2 $\uparrow$} \\
      \midrule
      No reference & 23.55 & 25.88 & 6.603 & 34.85 \\
      \midrule
      Perceptual loss & 23.42 & 25.98 & 6.455 & 34.90 \\
      \textbf{Ref. drift loss (ours)} &
      \textbf{23.49} & \textbf{26.22} & \textbf{6.485} & \textbf{35.07} \\
      \bottomrule
    \end{tabular}

    \captionof{table}{Effect of reference drift on Pick-a-Pic v2 test prompts.}
    \label{tab:reference-final}
  \end{minipage}
  \hfill
  \begin{minipage}[h]{0.45\textwidth}
    \centering
    \vspace{-5pt}
    \includegraphics[width=\linewidth]{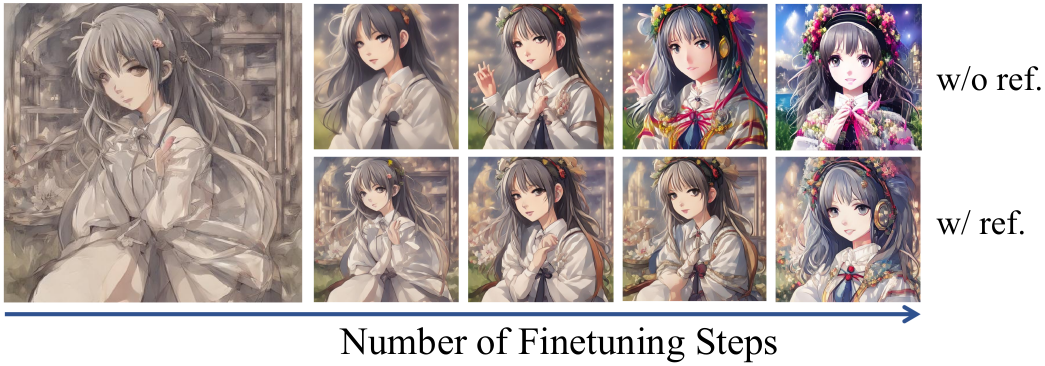}
    \vspace{-14.99pt}
    \captionof{figure}{Effect of the reference term during finetuning.}
    \label{fig:reg-train-step}
  \end{minipage}
\end{center}

\newpage
\noindent\textbf{Offline preference setting.} 
% Figure~\ref{fig:offline-drpo-dpo-pickscore} compares offline \DrPO{} with offline \DPOone{} variants using the same SD-Turbo base and Pick-a-Pic v2 test protocol. Here the positive and negative supports are fixed by the preference dataset rather than sampled online. \DrPO{} still improves over the base model, while the adapted \DPOone{} variants remain closer to SD-Turbo. This separates the effect of the feature-drift update from online candidate selection.
Figure~\ref{fig:offline-drpo-dpo-pickscore} compares a naive offline extension of \DrPO{} with offline \DPOone{} variants using the same SD-Turbo base and Pick-a-Pic v2 test protocol. Here $\mathcal{A}^+$ and $\mathcal{A}^-$ come from fixed preference pairs rather than online reward ranking. Compared to the naive \DPOone{} variant, offline \DrPO{} achieves better performance.

% \vspace{-4.5pt}
\vspace{-2pt}
\section{Discussions}
\vspace{-2pt}
% \vspace{-3pt}
\label{sec:discussion}

\textbf{Assumption on the population-rank reward model.} 
% The dipole reward model follows the Bradley--Terry formulation commonly assumed in RLHF methods, where preferences are modeled through reward differences between samples. However, it further treats each image in a preference batch symmetrically: positive samples define attraction and negative samples define repulsion under the same feature-space kernel. This assumption allows us to adapt gradient synthesis from Drifting models, but it also makes the representation critical. If the feature map does not encode the attribute being optimized, such as counting, layout, typography, or fine-grained identity, the resulting drift can be locally smooth but preference-misaligned.
% The population-rank reward model is compatible with the Bradley--Terry view used in RLHF, where pairwise preferences are induced by scalar reward differences. In \DrPO{}, this scalar reward is constructed non-parametrically from the current preference pairs: preferred samples define attraction and rejected samples define repulsion under the same feature-space kernel. This makes the feature map important. If $\phi$ does not encode the attribute being optimized, such as counting, layout, typography, or fine-grained identity, the resulting drift may be smooth in feature space but weakly aligned with the target preference.
The population-rank reward model is compatible with the Bradley--Terry view in RLHF, where pairwise preferences arise from scalar reward differences. In \DrPO{}, we construct a non-parametric scalar proxy as the log-ratio of kernel densities for preferred and rejected features. This ratio measures how winner-like a sample is within the current population, while its gradient attracts preferred samples and repels rejected ones. Thus, the feature map is important. If $\phi$ does not encode the optimized attribute, such as counting, layout, typography, or fine-grained identity, the resulting drift may be weakly aligned with the target preference.

\textbf{Challenges in offline finetuning.} 
% Although our experiments show that a naive extension of \DrPO{} to the offline setting outperforms a naive DPO baseline, its non-parametric synthesized reward gradients limit its ability to learn broad preferences and lead to sub-optimal performance. In the online setting, this gradient approximation is more reliable because the supports are sampled near the current model distribution. By contrast, the offline setting relies on off-policy samples and implicitly assumes that these local gradient estimates generalize beyond the data distribution. Moreover, offline preference datasets often contain only a single preference pair per text prompt, making the non-parametric estimate more fragile. Consequently, successful offline finetuning with \DrPO{} relies more heavily on the implicit regularization provided by the pretrained model and the latent feature representation.
Our offline experiment is a naive extension of the online update. In the online setting, preference pairs are sampled from the current generator, so the population-rank field is built near the distribution being updated. In the offline setting, the chosen and rejected samples can be off-policy and sparse, and many preference datasets contain only one pair per prompt. This makes the local non-parametric field more fragile. Better offline variants may need larger pair neighborhoods, prompt-conditioned retrieval, or an aggregated field over multiple preference pairs.

\textbf{Limitations and future work.}
% \DrPO{} is reward-model agnostic in interface, but not immune to reward misspecification. Optimizing a single automatic reward can change style, diversity, or prompt adherence in ways that the reward does not penalize. Stronger evaluation should include repeated runs, human preference studies, broader prompt distributions, and diagnostics for diversity and reference proximity. The current implementation also depends on candidate batch size and feature extraction, so training cost is shifted from reward backpropagation to candidate generation, reward ranking, and drift estimation.
\DrPO{} inherits the limitations of the target reward used for ranking. If a reward misses prompt attributes or encodes undesirable aesthetic tradeoffs, the resulting preference field can steer the generator in that direction. Training also still requires candidate generation, reward scoring, and feature extraction, so the computational advantage is largest when reward-model backpropagation is expensive or unavailable. The current implementation uses LoRA finetuning and finite candidate batches; scaling candidate selection, feature maps, and reward mixtures are useful directions for future work.

\vspace{-3pt}
\section{Concluding Remarks}
\vspace{-3pt}
\label{sec:conclusion}
% We presented \DrPO{}, a preference-finetuning method for one-step text-to-image generators that avoids reward backpropagation. \DrPO{} uses the target reward only to rank generated candidates, then builds a sample-based preference drift from positive and negative samples and a reference drift from the base generator. This yields a practical update direction for deterministic one-step models without policy likelihoods, denoising trajectories, or differentiating through the reward. Across generators and reward signals, \DrPO{} improves alignment efficiently, supports large or non-differentiable rewards, and shows promise for offline preference tuning.
% We presented \DrPO{}, a preference-finetuning method for one-step text-to-image generators that avoids reward backpropagation. \DrPO{} uses the target reward to rank on-policy samples, forms reward-ordered preference pairs, and converts them into a dipole drift in feature space. A reference drift from the frozen base generator stabilizes the update, and a detached feature regression trains the current generator along the resulting field. The method does not require policy likelihoods, denoising trajectories, or differentiating through the target reward, and inference remains a single generator call. Across one-step generators and reward signals, \DrPO{} improves reward-gradient-free preference finetuning, works with large and non-differentiable rewards, and gives an initial path toward offline preference tuning.
We presented \DrPO{}, a preference-finetuning method for one-step text-to-image generators that avoids reward backpropagation. \DrPO{} ranks on-policy samples with a target reward, converts reward-ordered pairs into a feature-space drift field, and stabilizes the update with a reference drift from the frozen base generator. Training requires neither policy likelihoods nor denoising trajectories and supports large or non-differentiable reward models. Empirical results show that \DrPO{} enables efficient reward-gradient-free finetuning and achieves higher preference win rates. Our initial experiments also suggest a promising path toward extending \DrPO{} to offline settings.

% \section{Acknowledgements and Disclosures}

\begin{ack}
Zhou Jiang thanks Haoquan Zhang and Ruotian Peng for the helpful discussion and support during this work.

\textbf{Funding.}
This work is supported by the National Natural Science Foundation of China (No. 62606411).

\textbf{Competing Interests.}
The authors declare no competing interests.
\end{ack}

\FloatBarrier

{
\footnotesize
\bibliographystyle{plainnat}
\bibliography{ref}
}

\appendix

\renewcommand{\thefigure}{A\arabic{figure}}
\renewcommand{\thetable}{A\arabic{table}}
\renewcommand{\thealgorithm}{A\arabic{algorithm}}
\renewcommand{\theHfigure}{A\arabic{figure}}
\providecommand{\theHtable}{}
\renewcommand{\theHtable}{A\arabic{table}}
\providecommand{\theHalgorithm}{}
\renewcommand{\theHalgorithm}{A\arabic{algorithm}}

\newpage
\appendix
\addcontentsline{toc}{section}{Appendix} %
\renewcommand \thepart{} %
\renewcommand \partname{}
\part{\vspace{-7mm}\Large{\centerline{Appendix}}\vspace{-2mm}}
\parttoc

\section{Construction of Drift Field}
\label{app:drift-implementation}

\begin{algorithm}[h]
\caption{Construction of Drift Field}
\label{alg:drift-field-impl}
\begin{algorithmic}[1]
\Require Query set $Z$, positive set $B^+$, negative set $B^-$, drift radii $\Omega$
\Function{DriftField}{$Z,B^+,B^-,\Omega$}
  \State Normalize query and support features by
  Eq.~\ref{eq:app-feature-scale}
  \State $V\gets 0$
  \For{$\rho\in\Omega$}
    \State Compute joint double-softmax affinities and split them into
    $A^{+,\rho}$ and $A^{-,\rho}$ by
    Eq.~\ref{eq:app-symmetric-affinity}
    \State Compute the cross-mass field $V^{(\rho)}$ by
    Eq.~\ref{eq:app-cross-mass-field}
    \State $V\gets V+\operatorname{RMSNorm}(V^{(\rho)})$
  \EndFor
  \State \Return $V$
\EndFunction
\end{algorithmic}
\end{algorithm}

% \paragraph{Batch construction.}
% Let $Z=\{z_i\}_{i=1}^{K}$ denote the current generated features. For the
% preference branch, the positive and negative support batches are
% \begin{equation}
% B_{\mathrm{pref}}^+=\mathcal{A}^+,\qquad
% B_{\mathrm{pref}}^-=\mathcal{A}^- .
% \end{equation}
% For the reference branch, they are
% \begin{equation}
% B_{\mathrm{ref}}^+=\mathcal{R},\qquad
% B_{\mathrm{ref}}^-=\operatorname{sg}(Z) .
% \end{equation}
% The detached current batch provides the current-model correction in
% $V_{\mathrm{ref}}$.

\paragraph{Feature normalization.}
Each call to $\Call{DriftField}{Z,B^+,B^-,\Omega}$ forms a support batch by concatenating the negative and positive support lists, $B=[B^-;B^+]$. If a sample appears in multiple preference pairs, we keep each occurrence in the
batch. Let $w_j$ denote the weight of support $b_j$. We use $w_j=1$ for all support entries in our experiments, so repeated entries contribute repeated support mass.

For each call, we compute distances using detached queries,
\begin{equation}
\Delta_{ij}=\|\operatorname{sg}(z_i)-b_j\|_2 ,
\end{equation}
and estimate a batch feature scale
\begin{equation}
s
=
\max\!\left(
\epsilon_s,\,
\frac{\operatorname{Mean}_{i,j}(w_j\Delta_{ij})}
{\operatorname{Mean}_{j}(w_j)}
\right).
\label{eq:app-feature-scale}
\end{equation}
Before constructing the drift field, we normalize feature coordinates by
\begin{equation}
s_z=\max\!\left(\epsilon_z,\,s/\sqrt{d}\right),\qquad
\widetilde{z}_i=z_i/s_z,\qquad
\widetilde{b}_j=b_j/s_z ,
\end{equation}
where $d$ is the feature dimension. This per-call normalization makes the estimator less sensitive to the absolute scale of the frozen feature.

\paragraph{Double-softmax affinity.}
For each radius $\rho$, let $D_{ij}=\Delta_{ij}/s$. By default, we use the Gaussian RBF logits
\begin{equation}
L_{ij}^{(\rho)}
=
-\frac{D_{ij}^2}{2\rho^2}.
\label{eq:app-rbf-logit}
\end{equation}
We compute affinities jointly over the concatenated support batch
$B=[B^-;B^+]$:
\begin{equation}
A_{ij}^{(\rho)}
=
\sqrt{
\operatorname{softmax}_{j}\!\left(L_{ij}^{(\rho)}\right)
\operatorname{softmax}_{i}\!\left(L_{ij}^{(\rho)}\right)
}\,w_j .
\label{eq:app-symmetric-affinity}
\end{equation}
The first softmax normalizes supports for each query, while the second normalizes queries for each support. Their geometric mean gives the double-softmax affinity. We split the resulting affinity matrix into $A^{+,\rho}$ and $A^{-,\rho}$ only after both softmax operations. We use the Gaussian RBF kernel by default and evaluate alternative radial kernels in our ablation.

\paragraph{Cross-mass drift.}
For each query, define the positive and negative affinity masses
\begin{equation}
m_i^{+,\rho}
=
\sum_{j\in B^+}A_{ij}^{+,\rho},
\qquad
m_i^{-,\rho}
=
\sum_{j\in B^-}A_{ij}^{-,\rho}.
\label{eq:app-cross-mass}
\end{equation}
The radius-specific drift field is
\begin{equation}
V_i^{(\rho)}
=
m_i^{-,\rho}
\sum_{j\in B^+}A_{ij}^{+,\rho}\widetilde{b}_j^+
-
m_i^{+,\rho}
\sum_{j\in B^-}A_{ij}^{-,\rho}\widetilde{b}_j^- .
\label{eq:app-cross-mass-field}
\end{equation}
Thus, positive supports attract a query in proportion to its negative mass, while negative supports repel it in proportion to its positive mass. This is the finite-batch cross-mass construction used for both drift branches.

\paragraph{RMS normalization and multi-radius aggregation.}
Each radius-specific field is normalized by its RMS magnitude over particles and feature coordinates,
\begin{equation}
\widehat{V}^{(\rho)}
=
\frac{V^{(\rho)}}{
\sqrt{
\max\!\left(
\operatorname{Mean}_{i,c}\left((V_{i,c}^{(\rho)})^2\right),
\epsilon_v
\right)
}
}.
\label{eq:app-drift-normalization}
\end{equation}
The final field is
\begin{equation}
V=\sum_{\rho\in\Omega}\widehat{V}^{(\rho)}.
\end{equation}
The detached target in Eq.~\ref{eq:drpo-target} is constructed from this aggregated field. Both the preference and reference branches use the same estimator, with different positive and negative support batches.

\section{Experimental Details}
\label{app:rewards}

\subsection{Target Reward Models}
\label{app:target-rewards}

\DrPO{} only requires the target reward to rank candidate images for the same prompt. We therefore instantiate the online preference construction with several reward backends, including PickScore, CLIP, AES, HPS-family rewards, ImageReward, and mixtures of these scores. Table~\ref{tab:reward-generality} shows that the optimized model moves toward the chosen target reward: PickScore training gives the largest PickScore gain, CLIP training gives the largest CLIP gain, HPS training gives the largest HPS gain, and ImageReward training gives the largest ImageReward gain.

\begin{table}[H]
  \centering
  \scriptsize
  \caption{Target-reward-conditioned online \DrPO{} on Pick-a-Pic v2 test prompts. Entries except AES and ImageReward are multiplied by $100$.}
  \label{tab:reward-generality}
  \setlength{\tabcolsep}{3.4pt}
  \renewcommand{\arraystretch}{0.98}
  % \resizebox{0.8\textwidth}{!}{%
  \begin{tabular}{lccccc}
    \toprule
    Target reward model &
    PickScore $\uparrow$ &
    CLIP $\uparrow$ &
    AES $\uparrow$ &
    IR $\uparrow$ &
    HPSv2 $\uparrow$ \\
    \midrule
    \rowcolor{gray!15} SD-Turbo & 21.88 & 26.47 & 6.054 & 5.75 & 35.80 \\
    PickScore & \textbf{23.40} & 26.28 & 6.390 & 10.09 & 35.24 \\
    CLIP & 21.04 & \textbf{28.62} & 5.491 & 3.31 & 36.82 \\
    HPS & 21.13 & 27.82 & 5.396 & 5.60 & \textbf{39.42} \\
    IR & 21.54 & 23.98 & 6.314 & \textbf{16.25} & 32.56 \\
    CLIP+PickScore & 23.01 & 28.60 & 6.328 & 10.08 & 37.36 \\
    AES+PickScore & 22.78 & 23.61 & \textbf{7.429} & 8.51 & 31.38 \\
    CLIP+AES+PickScore & 22.94 & 27.18 & 7.064 & 10.70 & 35.08 \\
    \bottomrule
  \end{tabular}
  % }
\end{table}

\subsection{Data and Evaluation Sets}
\label{app:data}

\paragraph{Online training.}
For online \DrPO{}, each training example contains only a prompt. At each update, the current generator samples a candidate batch for the prompt, the target reward ranks the candidates, and the ordered candidates are used to form preference pairs. Each online method is trained for the same optimization steps with the same prompt-sampling protocol. 

\paragraph{Offline preference data.}
For the offline variant, each record contains a prompt and one or more chosen/rejected image pairs. The chosen and rejected images replace the online ranked candidates when constructing the preference pairs; the reference drift and training loss are unchanged.

% \paragraph{GenEval.}
% For GenEval experiments, prompts are paired with structured compositional
% constraints such as object category, count, color, position, and attribute
% binding. The GenEval evaluator returns task-level correctness scores, which are
% used only to rank candidate images during online training. For the GenEval training set, we train separately on each subtask and evaluate on the corresponding test prompts. Training prompts are generated from the corresponding task templates, with all test prompts removed to prevent overlap.
\paragraph{GenEval.}
In the GenEval experiments, prompts are paired with structured compositional constraints, including object category, count, color, position, and attribute binding. The GenEval evaluator provides task-level correctness scores, which are used solely for ranking candidate images during online training. 
% For the GenEval training set, we train each subtask independently and evaluate on the corresponding test prompts. Training prompts are generated from the respective task templates, with all test prompts excluded to prevent any overlap between training and evaluation.

\paragraph{Evaluation.}
The main scalar-metric evaluations use the Pick-a-Pic v2 test prompts. Unless otherwise stated, one-step models are evaluated with deterministic sampling and guidance scale $0.0$. We report one image per prompt for PickScore, CLIP, AES, HPSv2, HPSv3, and ImageReward; standard errors are estimated by bootstrapping prompts.

\subsection{Sampling and Reproducibility}
\label{app:reproducibility}

Evaluation supports three sampling modes: the original SD-Turbo checkpoint, a finetuned U-Net checkpoint, and a LoRA adapter. For structured compositional benchmarks, the same harness reads the benchmark metadata and runs the corresponding evaluator. All paper-facing comparisons fix prompt sets, random seeds, sampling configuration, and the distinction between target rewards used
for training and metrics used only for evaluation.

\subsection{Qwen3-VL Pairwise Evaluation Prompt}
\label{app:qwen3vl-prompt}

For the pairwise preference results in Figure~\ref{fig:qwen3vl_drpo_preference_by_dataset}, we use Qwen3-VL~\citep{qwen2025qwen3vl} as a strict text-to-image judge. Each comparison shows two images generated from the same prompt, and the judge returns a JSON decision with ties allowed. The prompt template is:

\begin{Verbatim}[fontsize=\scriptsize,breaklines=true,frame=single,framesep=2mm]
You are a strict text-to-image judge comparing two generated images.

You will see Image A and Image B. They were generated from the same text prompt.

Text prompt:
"""{prompt}"""

Choose the image that is better as a clean, coherent realization of the prompt.

Rubric, in order:
1. Semantic fidelity: prefer the image where the prompt's main subject is immediately recognizable and the requested attributes, relations, actions, and scene are correctly expressed.
2. Global coherence: prefer one integrated scene over a collage-like, over-stylized, or visually confused image. Objects should have plausible shape, scale, pose, and spatial relation.
3. Artifact avoidance: strongly penalize malformed bodies, fused objects, duplicated parts, warped geometry, messy textures, unreadable required text, and unnatural distortions.
4. Aesthetic quality: after the above, prefer tasteful lighting, color harmony, composition, and detail. Do not choose an image only because it is sharper, glossier, more saturated, or more dramatic.

Decision rule:
- Major semantic or structural errors should usually lose, even if the image looks polished.
- If both satisfy the text similarly, choose the more coherent and artifact-free image.
- If both are genuinely comparable, answer tie.
- Do not prefer A or B because of position or label.
- Return only valid JSON.

Output schema:
{"winner":"A"|"B"|"tie","confidence":0.0-1.0,"reason":"12 words or fewer"}
\end{Verbatim}

\section{Baselines and Ablations}
\label{app:baselines}

This section describes the one-step preference baselines used in our comparisons. All baselines use the same prompt sets, one-step sampler, and evaluation suite as for \DrPO{}.

\paragraph{\PSO{} baseline.}
\PSO{} is a preference-finetuning method for timestep-distilled diffusion models~\citep{miao2024pso}. It optimizes pairwise preferences by increasing the relative margin between preferred images and reference images sampled from the distilled model. We include it as a native pairwise baseline for one-step distilled generators.

\subsection{\DRaFT{} Reward-Gradient Baseline}
\label{app:draft}

\DRaFT{} directly optimizes a differentiable image reward through the generator and decoder~\citep{clark2024draft}. In our one-step setting, the U-Net predicts a clean latent
\[
\mu_\theta=T(z,f_\theta(z,t,c)),
\]
which is decoded and scored by a differentiable reward model $R$. We use the same one-step sampler, LoRA parameterization, prompt batches, and evaluation protocol as \DrPO{}. The training objective is
\[
\mathcal{L}_{\text{DRaFT}}
=
-\frac{1}{K}\sum_{k=1}^{K} R(D(\mu_{\theta,k}),p)
+\lambda_{\mathrm{ref}}
\frac{1}{K}\sum_{k=1}^{K}
\|\mu_{\theta,k}-\mu_{\mathrm{ref},k}\|_2^2,
\]
where $D$ is the VAE decoder and $\mu_{\mathrm{ref},k}$ is the frozen base-model prediction for the same prompt and noise. This baseline represents the standard reward-backpropagation regime: it can exploit the local gradient of a differentiable reward, but it requires storing and backpropagating through the reward network. For SD-Turbo and SDXL-Turbo, the implementation evaluates generated samples in chunks to keep the effective generated batch size matched to \DrPO{} while controlling memory.

\subsection{\VGGFlow{} Reward-Gradient Target Baseline}
\label{app:vggflow}

\VGGFlow{} adapts value-gradient guidance for flow matching alignment to the one-step latent prediction setting~\citep{liu2025value}. Rather than directly maximizing the scalar reward as in \DRaFT{}, it first differentiates the reward with respect to the generated clean latent and then constructs a target latent around the frozen reference prediction:
\[
\mu^{\mathrm{target}}_k
=
\mu_{\mathrm{ref},k}
+ \eta(t)\lambda_{\mathrm{r}}
\operatorname{clip}\!\left(
\nabla_{\mu_{\theta,k}} R(D(\mu_{\theta,k}),p)
\right).
\]
The student is trained by latent target matching,
\[
\mathcal{L}_{\VGGFlow}
=
\frac{1}{K}\sum_{k=1}^{K}
\|\mu_{\theta,k}-\mu^{\mathrm{target}}_k\|_2^2
+\lambda_{\mathrm{ref}}
\frac{1}{K}\sum_{k=1}^{K}
\|\mu_{\theta,k}-\mu_{\mathrm{ref},k}\|_2^2.
\]
Here $\eta(t)$ is fixed in the one-step experiments because the generation timestep is fixed, and the reward gradient is norm-clipped before constructing the target. This implementation preserves the one-step deployment interface but keeps the defining reward-gradient path of VGG-Flow. We omit the multi-step value-network consistency branch from the original flow-matching setting, since SD-Turbo and SDXL-Turbo expose only a single distilled transition in our experiments.

\subsection{\DPOone{} Pairwise Baselines}
\label{app:dpo}

We adapt DPO to one-step generation by defining the preference margin on the clean latent predicted from a fixed noisy latent and prompt. For text embedding $c$, timestep $t=999$, and initial noise $z$, the student and reference U-Nets produce
\[
\mu_{\theta}=T(z,f_\theta(z,t,c)), \qquad
\mu_{\mathrm{ref}}=T(z,f_{\mathrm{ref}}(z,t,c)),
\]
where $T(\cdot)$ is the deterministic SD-Turbo one-step conversion. Given chosen and rejected targets $y^+$ and $y^-$ mapped into a preference space $\psi$, we define
\[
g_\theta =
s\!\left(\psi(\mu_\theta), \psi(y^+)\right)
-
s\!\left(\psi(\mu_\theta), \psi(y^-)\right),
\]
and define $g_{\mathrm{ref}}$ analogously using $\mu_{\mathrm{ref}}$. The loss is
\[
\mathcal{L}_{\DPOonemath}
=
-\log \sigma\!\left(\beta[g_\theta-g_{\mathrm{ref}}]\right)
+
\lambda_{\mathrm{ref}}\|\mu_\theta-\mu_{\mathrm{ref}}\|_2^2.
\]
We use $\beta=30$ and $\lambda_{\mathrm{ref}}=0$ in the reported runs. We test two choices of $\psi$: raw VAE latent space and frozen MAE-latent feature space.

\begin{algorithm}[t]
\caption{\DPOone{} baseline}
\label{alg:dpo-one}
\begin{algorithmic}[1]
\Require Student U-Net $f_\theta$, frozen reference $f_{\mathrm{ref}}$, offline triples $(p,y^+,y^-)$, text encoder $E_{\mathrm{text}}$, one-step transform $T$, preference map $\psi$, inverse temperature $\beta$
\For{each optimization step}
  \State Sample $(p,y^+,y^-)$ and noise latent $z$; set $c=E_{\mathrm{text}}(p)$
  \State $\mu_\theta\gets T(z,f_\theta(z,t,c))$, \quad $\mu_{\mathrm{ref}}\gets T(z,f_{\mathrm{ref}}(z,t,c))$
  \State Compute $h_\theta=\psi(\mu_\theta)$, $h_{\mathrm{ref}}=\psi(\mu_{\mathrm{ref}})$, $h^+=\psi(y^+)$, $h^-=\psi(y^-)$
  \State Define $s(a,b)=-\|a-b\|_2^2$
  \State $g_\theta\gets s(h_\theta,h^+)-s(h_\theta,h^-)$
  \State $g_{\mathrm{ref}}\gets s(h_{\mathrm{ref}},h^+)-s(h_{\mathrm{ref}},h^-)$
  \State $\mathcal{L}\gets-\log\sigma(\beta(g_\theta-g_{\mathrm{ref}}))+\lambda_{\mathrm{ref}}\|\mu_\theta-\mu_{\mathrm{ref}}\|_2^2$
  \State Update $\theta$ by descending $\nabla_\theta\mathcal{L}$
\EndFor
\end{algorithmic}
\end{algorithm}

\FloatBarrier

\subsection{\GRPOone{} Policy-Gradient Baseline}
\label{app:flowgrpo}

We include \GRPOone{} as a policy-gradient-style comparison for one-step generators. Since a one-step model has no denoising trajectory, the policy action is defined in clean latent space. For a prompt embedding $c$ and a group of initial noises $\{z_k\}_{k=1}^{K}$,
\[
\mu_{\theta,k}=T(z_k,f_\theta(z_k,t,c)),\qquad
\mu_{\mathrm{ref},k}=T(z_k,f_{\mathrm{ref}}(z_k,t,c)).
\]
The action is a perturbed clean latent,
\[
a_k=\operatorname{sg}(\mu_{\theta,k}^{\mathrm{old}})+\sigma_a u_k,
\qquad
u_k=\frac{z_k-\operatorname{mean}(z_k)}
{\operatorname{std}(z_k)+\epsilon}.
\]
After decoding $a_k$ and scoring the image, rewards are normalized within the same prompt group:
\[
A_k=
\operatorname{clip}
\left(
\alpha_{\mathrm{adv}}
\frac{r_k-\operatorname{mean}(r)}{\operatorname{std}(r)+\epsilon_r},
-A_{\max},A_{\max}
\right).
\]
The current policy mean is recomputed and assigned a Gaussian-style log probability
\[
\ell_\theta(a_k)
=
-\frac{1}{2}
\operatorname{mean}
\left[
\left(
\frac{a_k-\mu_{\theta,k}}{\sigma_a}
\right)^2
\right],
\]
where the mean is over latent dimensions. With
\[
\rho_k=\exp(\operatorname{clip}(\ell_\theta(a_k)-\ell_{\mathrm{old}}(a_k),-M,M)),
\]
we use the clipped policy-gradient objective
\[
\mathcal{L}_{\mathrm{policy}}
=
-\frac{1}{K}\sum_{k=1}^{K}
\min
\left(
\rho_k A_k,\;
\operatorname{clip}(\rho_k,1-\epsilon,1+\epsilon)A_k
\right).
\]
The total loss is
\[
\mathcal{L}
=
\mathcal{L}_{\mathrm{policy}}
+
\lambda_{\mathrm{ref}}
\frac{1}{K}\sum_{k=1}^{K}\|\mu_{\theta,k}-\mu_{\mathrm{ref},k}\|_2^2
+
\beta_{\mathrm{kl}}
\frac{1}{K}\sum_{k=1}^{K}(\rho_k-1-\log\rho_k).
\]
In the reported setting, $\sigma_a=0.05$, $\epsilon=0.2$, $M=10$,
$\alpha_{\mathrm{adv}}=1$, $A_{\max}=5$, $\lambda_{\mathrm{ref}}=0.05$, and
$\beta_{\mathrm{kl}}=0$.

\begin{algorithm}[t]
\caption{\GRPOone{} baseline}
\label{alg:flowgrpo}
\begin{algorithmic}[1]
\Require Student U-Net $f_\theta$, frozen reference $f_{\mathrm{ref}}$, reward model $R$, text encoder $E_{\mathrm{text}}$, decoder $D$, one-step transform $T$, group size $K$
\For{each prompt}
  \State $c\gets E_{\mathrm{text}}(p)$; sample $\{z_k\}_{k=1}^{K}$
  \For{$k=1,\dots,K$}
    \State $\mu_k^{\mathrm{old}}\gets T(z_k,f_\theta(z_k,t,c))$ without gradient
    \State $a_k\gets\operatorname{sg}(\mu_k^{\mathrm{old}})+\sigma_a\,\operatorname{normalize}(z_k)$
    \State $x_k\gets D(a_k)$; \quad $r_k\gets R(x_k,c)$
  \EndFor
  \State Normalize rewards within the prompt group to obtain $\{A_k\}$
  \For{$k=1,\dots,K$}
    \State Recompute $\mu_k\gets T(z_k,f_\theta(z_k,t,c))$ and $\mu_k^{\mathrm{ref}}\gets T(z_k,f_{\mathrm{ref}}(z_k,t,c))$
    \State Compute $\rho_k$ from the Gaussian log-probability ratio of $a_k$
  \EndFor
  \State Update $\theta$ with the clipped policy-gradient loss and reference latent anchor
\EndFor
\end{algorithmic}
\end{algorithm}

\FloatBarrier
\clearpage

\section{Feature Representations for DrPO}
\label{app:representation}

\begin{wraptable}{r}{0.55\textwidth}
\vspace{-0.8em}
\centering
\caption{Comparison of feature representations for \DrPO{} on SDXL-Turbo with 100 training steps.}
\label{tab:sdxl-reference-features}
\footnotesize
\setlength{\tabcolsep}{4pt}
\begin{tabular}{lcc}
\toprule
Feature representation & PickScore $\uparrow$ & Gain \\
\midrule
SDXL-Turbo & 22.449 & -- \\
VAE + MAE & 22.924 & +0.475 \\
VAE + DINOv2 & 22.869 & +0.420 \\
\textbf{Reference U-Net} & \textbf{23.111} & \textbf{+0.662} \\
\bottomrule
\end{tabular}
\vspace{-0.8em}
\end{wraptable}

The effectiveness of drifting methods depends on the feature representation. Beyond external pretrained encoders, recent work has shown that diffusion models can provide effective representations through their intermediate features~\citep{zhang2026teacherfeature}. Motivated by this finding, we further explore using the hidden states of the frozen reference U-Net directly as the feature representation for \DrPO{}.

Table~\ref{tab:sdxl-reference-features} compares three representations on SDXL-Turbo under a matched 100-step training budget. All improve PickScore over the base model, with reference U-Net features achieving the largest gain (+0.662). These results demonstrate that \DrPO{} can effectively reuse the generator's pretrained representations without requiring an additional feature encoder.

% \section{Feature Representations for DrPO}
% \label{app:representation}

% The effectiveness of drifting methods depends on the feature representation. Beyond external pretrained encoders, recent work has shown that diffusion models can provide effective representations through their intermediate features~\citep{zhang2026teacherfeature}. Motivated by this finding, we further explore using the hidden state of the frozen ref model directly as the feature extractor for \DrPO{}.

% Table~\ref{tab:sdxl-reference-features} compares three representations on SDXL-Turbo under a matched 100-step training budget. All improve PickScore over the base model, with reference U-Net features achieving the largest gain (+0.662). These results demonstrate that \DrPO{} can effectively reuse the generator's pretrained representations without requiring an additional feature encoder.

% \begin{table}[t]
% \centering
% % \small
% \caption{Comparison of feature representations for \DrPO{} on SDXL-Turbo with 100 training steps.}
% \label{tab:sdxl-reference-features}
% \setlength{\tabcolsep}{9pt}
% % \renewcommand{\arraystretch}{1.1}
% \begin{tabular}{lcc}
% \toprule
% Feature representation & PickScore $\uparrow$ & Gain \\
% \midrule
% SDXL-Turbo & 22.449 & -- \\
% VAE + MAE & 22.924 & +0.475 \\
% VAE + DINOv2 & 22.869 & +0.420 \\
% \textbf{Reference U-Net} & \textbf{23.111} & \textbf{+0.662} \\
% \bottomrule
% \end{tabular}
% \end{table}

\section{Additional Qualitative Results}
\label{app:additional-qualitative}

Figures~\ref{fig:qualitative-comparison-full} and
\ref{fig:reward-model-matrix-full} provide the full qualitative grids corresponding to the compact examples in the main text. All samples use matched seed $42$.

\paragraph{Reward-model prompts.}
The compact reward-model matrix in Figure~\ref{fig:reward-selector-matrix} uses four Pick-a-Pic v2 test prompts: ``Cyberpunk cat''; ``astronaut in space with a two handed sword in plate armor in front of the earth''; ``An mystical owl sitting on a tree branch in a magical Forest, art style of nicoletta ceccoli''; and ``Gothic cathedral in a stormy night''. The full matrix in Figure~\ref{fig:reward-model-matrix-full} additionally includes ``smily french fries'' and ``Raindrop, macro photograph, colorful, reflections, HD, 4k''.

\setcounter{figure}{2}

\begin{figure}[H]
  \centering
  \includegraphics[width=0.95\textwidth]{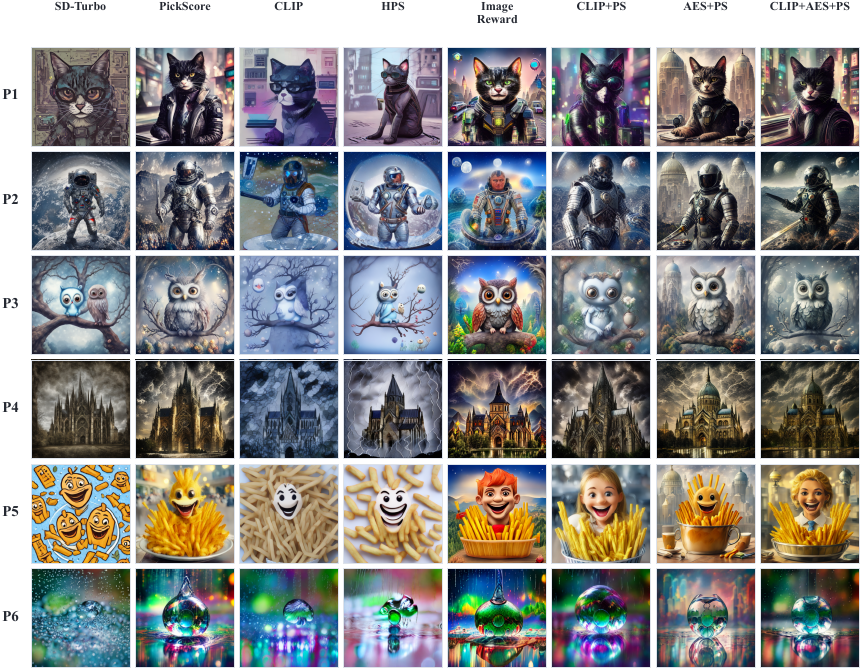}
  \caption{Full reward-model qualitative matrix corresponding to Figure~\ref{fig:reward-selector-matrix}. Rows are prompts, columns are target reward models, and all samples use matched seed $42$.}
  \label{fig:reward-model-matrix-full}
\end{figure}
\clearpage

\begin{figure}[H]
  \centering
  \includegraphics[width=0.95\textwidth]{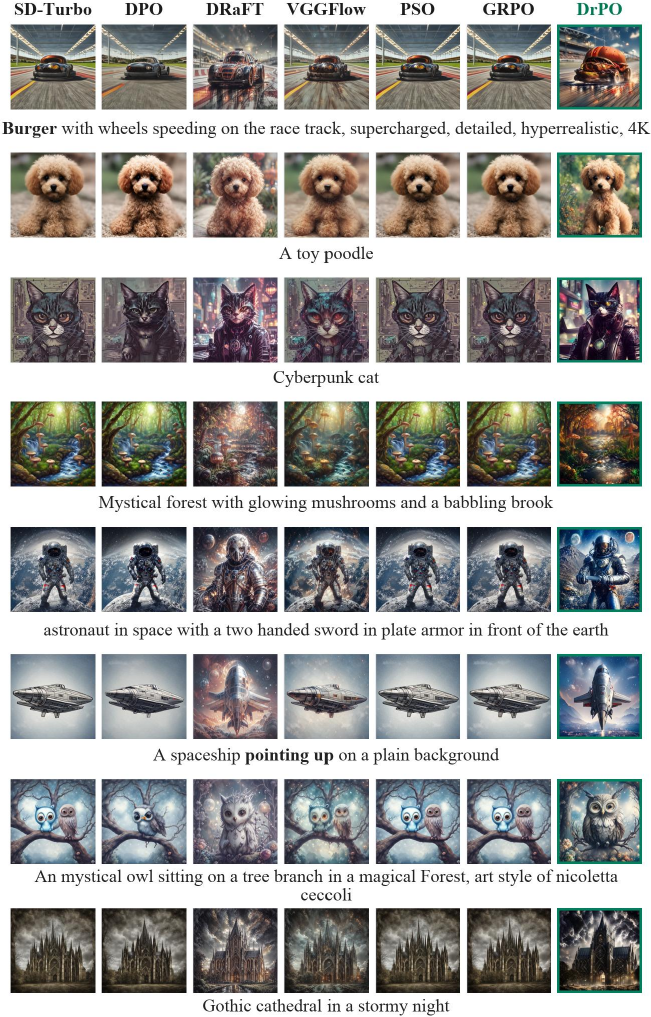}
  \caption{Full qualitative comparison corresponding to Figure~\ref{fig:qualitative-comparison}. The grid uses a larger selected prompt set with matched seed $42$.}
  \label{fig:qualitative-comparison-full}
\end{figure}

\begin{figure}[H]
\centering
\includegraphics[width=0.995\textwidth]{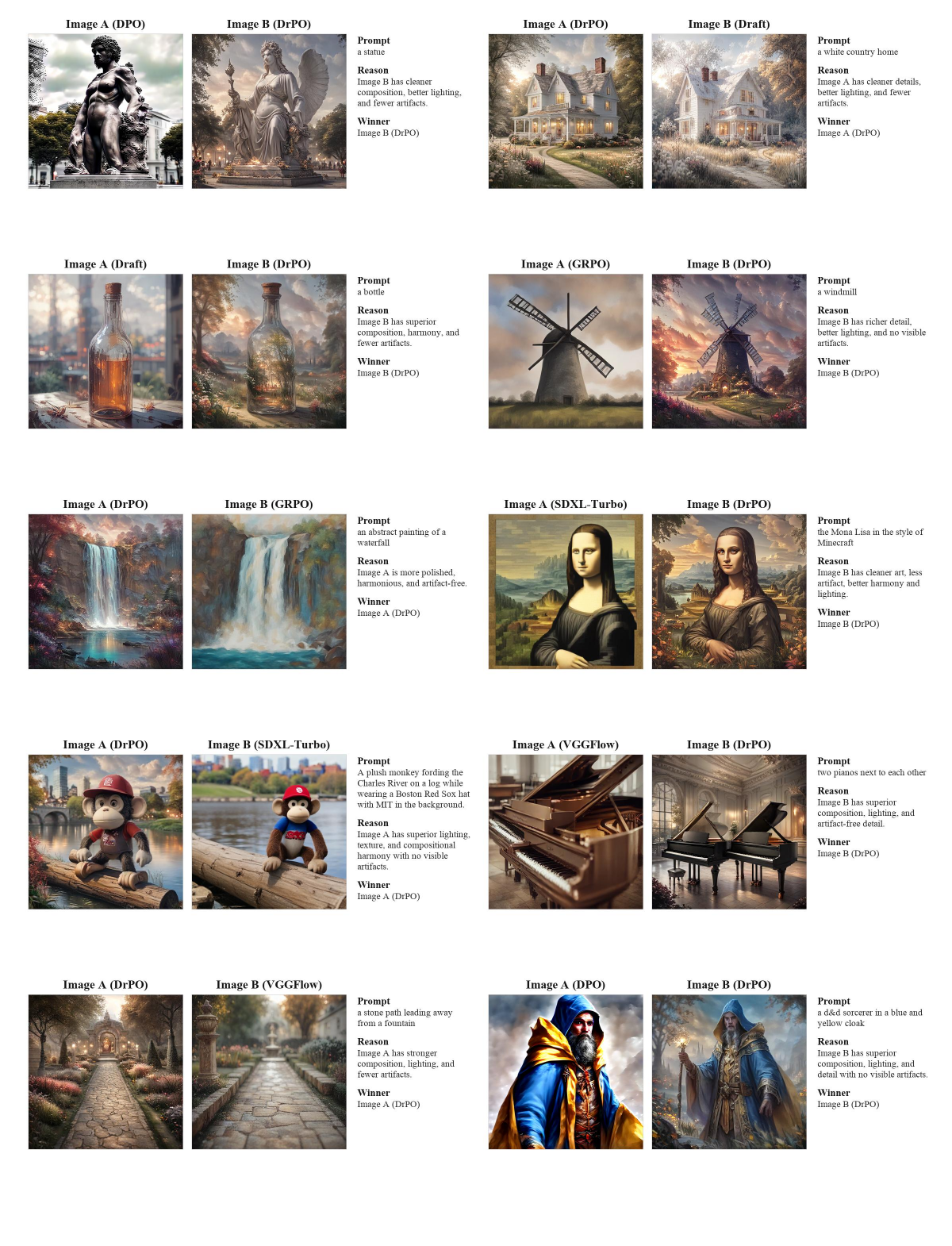}
\caption{Pairwise examples from the Qwen3-VL presentation-quality evaluation. Each row contains two matched-seed pairs.}
\label{fig:appendix-pairwise-cards-qwen3vl-selected}
\end{figure}

\begin{figure}[H]
\centering
\includegraphics[width=0.995\textwidth]{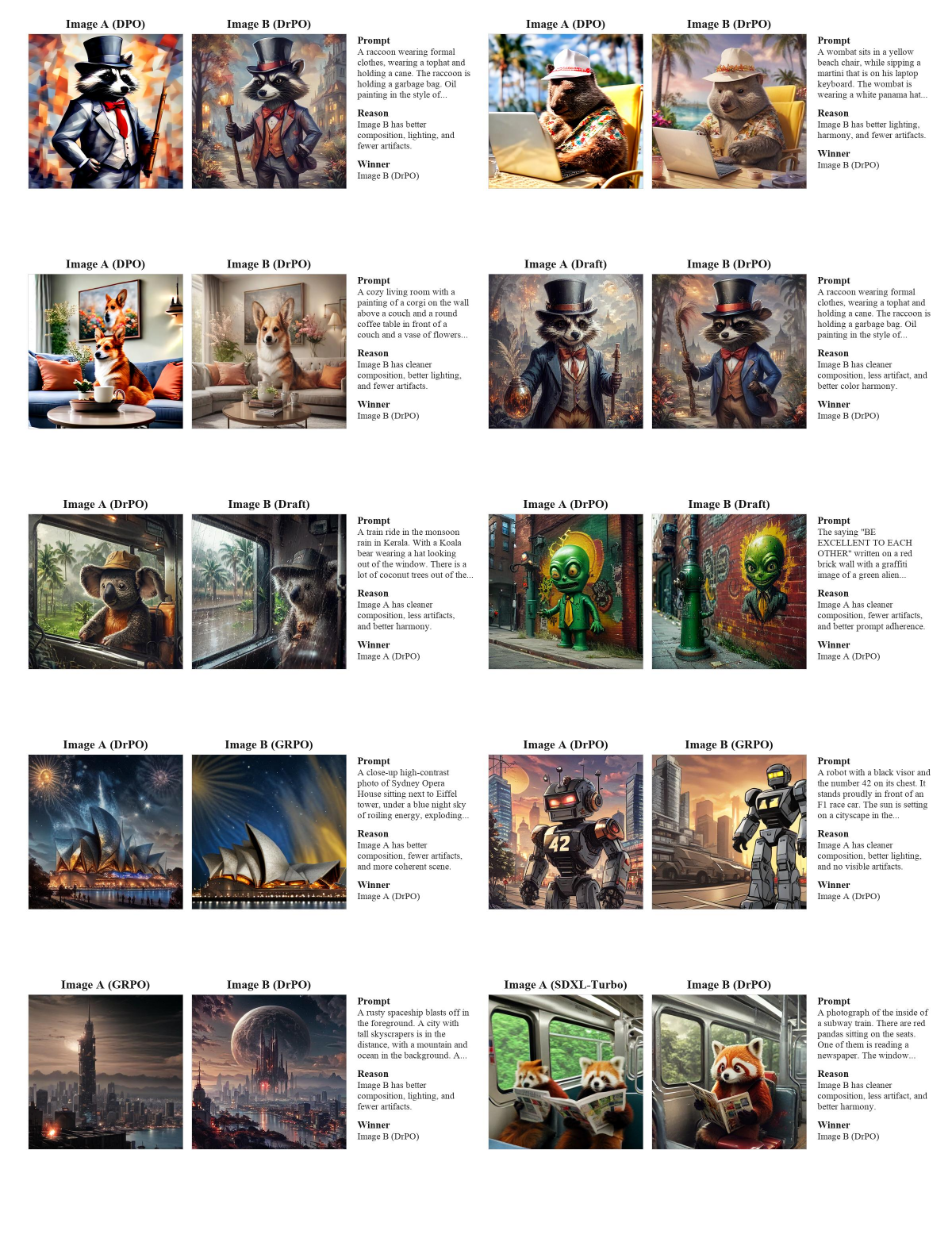}
\caption{Additional pairwise examples from the Qwen3-VL presentation-quality evaluation.}
\label{fig:appendix-pairwise-cards-qwen3vl}
\end{figure}

\begin{figure}[H]
\centering
\includegraphics[width=0.995\textwidth]{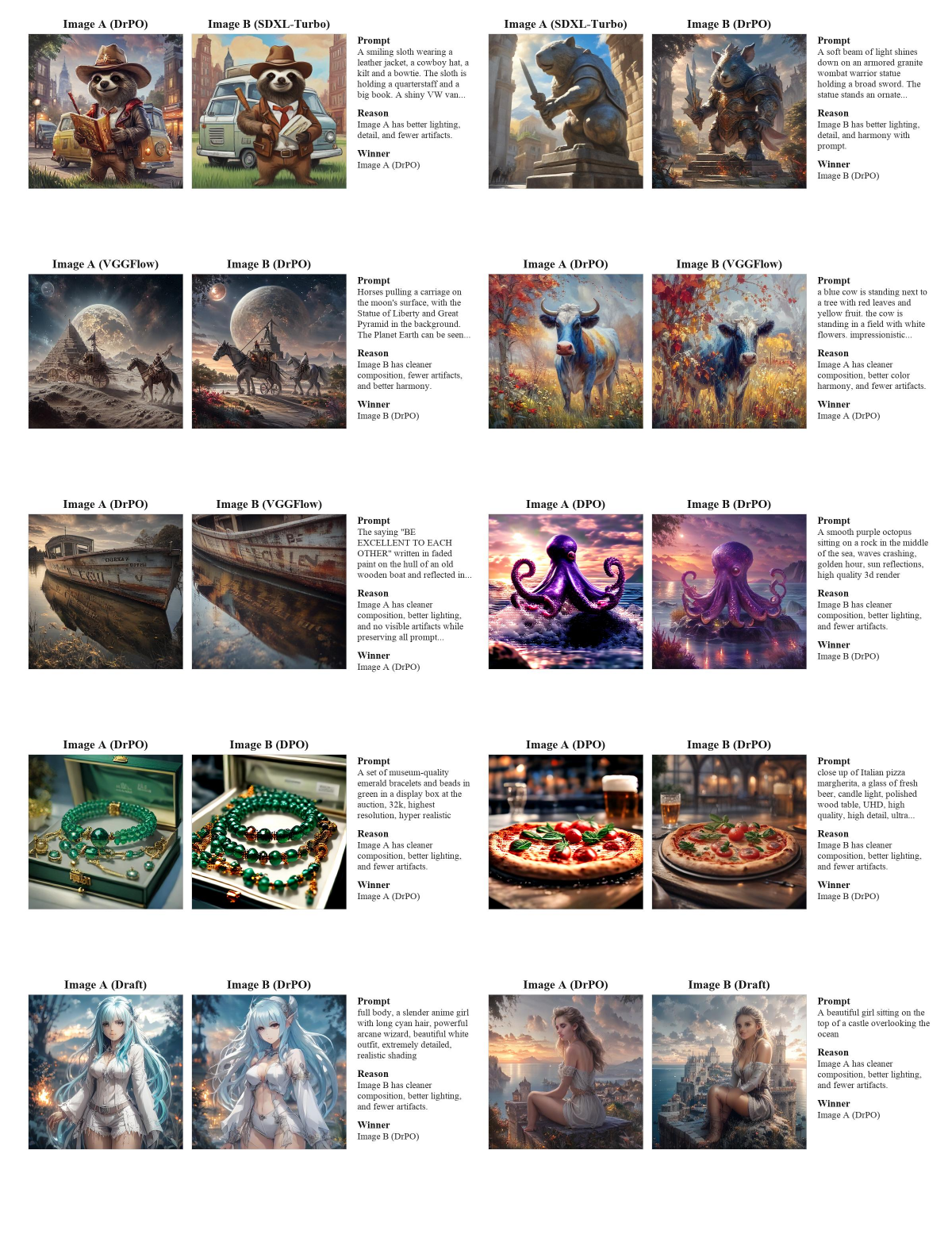}
\caption{Additional pairwise examples from the Qwen3-VL presentation-quality evaluation.}
\end{figure}

\begin{figure}[H]
\centering
\includegraphics[width=0.995\textwidth]{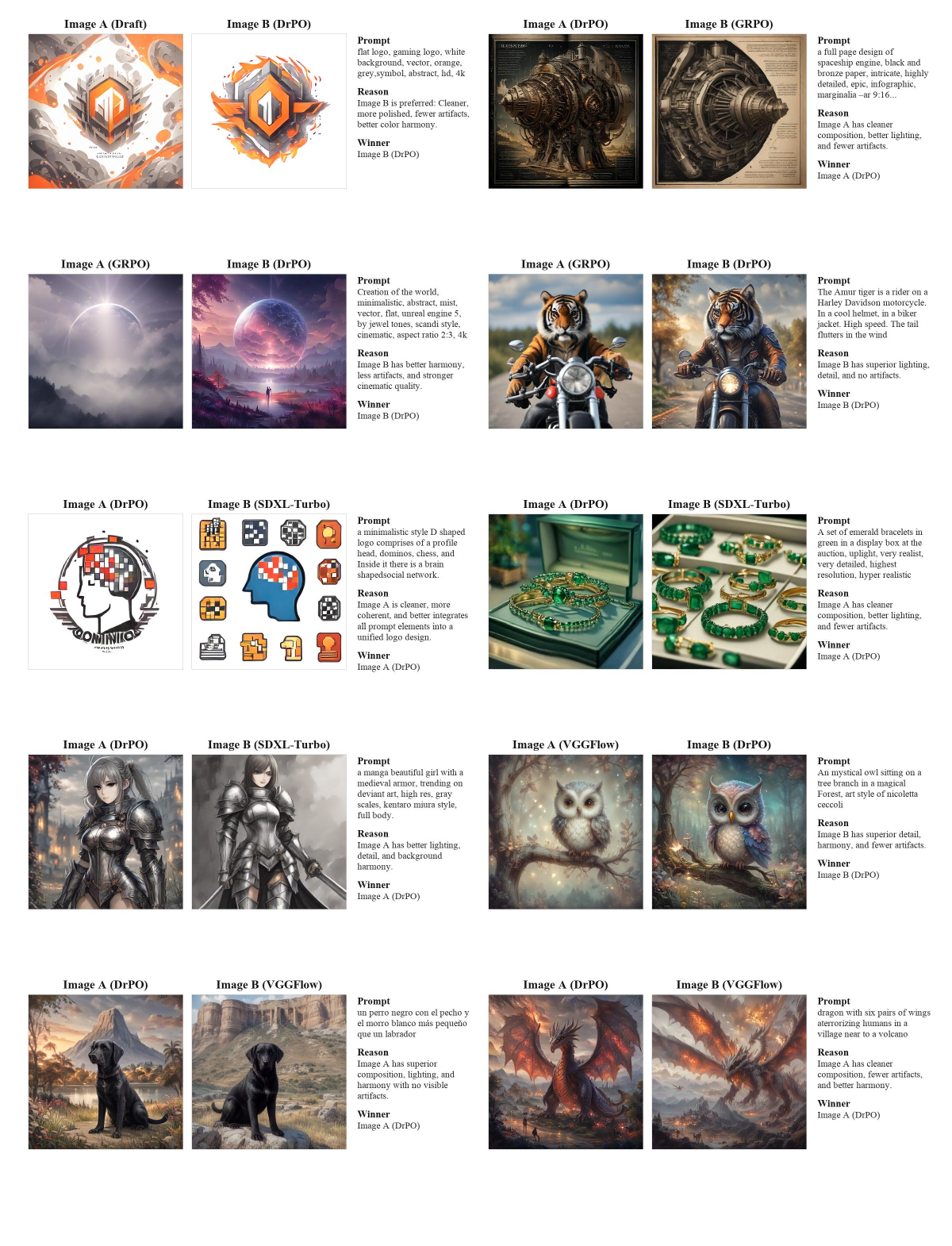}
\caption{Additional pairwise examples from the Qwen3-VL presentation-quality evaluation.}
\end{figure}

% \newpage
% \input{checklist.tex}

\end{document}